\documentclass[pdflatex,sn-mathphys-num]{sn-jnl}

\usepackage{graphicx}%
\usepackage{multirow}%
\usepackage{amsmath,amssymb,amsfonts}%
\usepackage{amsthm}%
\usepackage{mathrsfs}%
\usepackage[title]{appendix}%
\usepackage{xcolor}%
\usepackage{textcomp}%
\usepackage{manyfoot}%
\usepackage{booktabs}%
\usepackage{algorithm}
\usepackage{algpseudocode}
\usepackage{listings}%
\usepackage{subfigure}
\usepackage{subcaption}
\usepackage{placeins}
\usepackage{tabularx}  
\theoremstyle{thmstyleone}%
\theoremstyle{thmstyletwo}%
\theoremstyle{thmstylethree}%
\begin{document}

\title[CloudBreaker: Breaking the cloud covers of Sentinel-2 Images]{CloudBreaker: Breaking the Cloud Covers of Sentinel-2 Images using Multi-Stage Trained Conditional Flow Matching on Sentinel-1}




\author[1]{\fnm{Saleh Sakib} \sur{Ahmed}}\email{birdhunterx91@gmail.com}


\author*[2]{\fnm{Sara} \sur{Nowreen}}\email{snowreen@iwfm.buet.ac.bd}

\author*[1]{\fnm{M. Sohel} \sur{Rahman}}\email{msrahman@cse.buet.ac.bd}

\affil[1]{\orgdiv{Computer Science and Engineering}, \orgname{Bangladesh University of Engineering and Technology}, \orgaddress{\street{Palashi}, \city{Dhaka}, \country{Bangladesh}}}

\affil[2]{\orgdiv{Institute of Water and Flood Management  }, \orgname{Bangladesh University of Engineering and Technology}, \orgaddress{\street{Palashi}, \city{Dhaka}, \country{Bangladesh}}}



\abstract{Cloud cover and nighttime conditions remain significant limitations in satellite-based remote sensing, often restricting the availability and usability of multi-spectral imagery. In contrast, Sentinel-1 radar images are unaffected by cloud cover and can provide consistent data regardless of weather or lighting conditions. To address the challenges of limited satellite imagery, we propose CloudBreaker, a novel framework that generates high-quality multi-spectral Sentinel-2 signals from Sentinel-1 data. This includes the reconstruction of optical (RGB) images as well as critical vegetation and water indices such as NDVI and NDWI.We employed a novel multi-stage training approach based on conditional latent flow matching and, to the best of our knowledge, are the first to integrate cosine scheduling with flow matching. CloudBreaker demonstrates strong performance, achieving a Fréchet Inception Distance (FID) score of 0.7432, indicating high fidelity and realism in the generated optical imagery. The model also achieved Structural Similarity Index Measure (SSIM) of 0.6156 for NDWI and 0.6874 for NDVI, indicating a high degree of structural similarity. This establishes CloudBreaker as a promising solution for a wide range of remote sensing applications where multi-spectral data is typically unavailable or unreliable.}




\maketitle

\section{Introduction}\label{sec:intro}

Cloud cover poses significant challenges in satellite-based Earth observation. 
Persistent cloudiness hampers the acquisition of clear satellite images, affecting various applications. Sentinel-2, part of the European Space Agency's Copernicus Programme \cite{copernicus_sentinel2}, has been providing high-resolution multispectral imagery, which includes additional signals, such as near-infrared (NIR) in addition to standard Optical (RGB) images. Using these additional signals, we can derive various important indices, such as the Normalized Difference Water Index (NDWI) and the Normalized Difference Vegetation Index (NDVI). These signals have applications in agriculture, forestry, water quality monitoring, disaster response, urban planning, climate studies, and even in some military applications \cite{sentinel2}. RGB images are used for various applications, from urban mapping of peaceful cities \cite{corbane2008rapid} to detecting military targets \cite{bandarupally2020detection} and conducting reconnaissance operations \cite{wang2014military}. NDVI is used for civilian purposes, such as crop yield estimation \cite{quarmby1993use} and vegetation health monitoring \cite{bento2020roles}, as well as for military applications, such as detecting war damage \cite{shelestov2023war} and using field spectroscopy for defence and security (e.g., locating underground structures) \cite{melillos2016integrated}. NDWI is also used in war damage analysis \cite{vlasova2023monitoring,skybamon25}, and for critical tasks, such as detection of water bodies \cite{ozelkan2020water} and flood monitoring \cite{albertini2022detection}. These signals hold great importance for both civilian and military applications. However, the overcast cloud barrier limits our ability to use these valuable signals. In contrast, Sentinel-1 \cite{sentinel1} operates using Synthetic Aperture Radar (SAR), which is immune to cloud cover and lighting conditions, making it a promising alternative data source.

Some attempts to overcome this `cloud-barrier' have been reported in the literature. Kim et al. \cite{kim2024conditional} used a modified version of the Brownian Bridge Diffusion model \cite{li2023bbdm} to generate Optical (RGB) images from Very-High-Resolution (VHR) SAR using the MSAW dataset \cite{shermeyer2020spacenet}, which contains pairs of optical and SAR images. Their model introduced the concept of adding the initial SAR image at every step of the diffusion process, effectively conditioning the process on the initial image. 
Ahmed et al. \cite{sakib2025light} on the other hand used a U-Net architecture \cite{ronneberger2015u} to generate NDWI from Sentinel-1 signals. 
However, neither approach fully utilized the full potential of generating multispectral imagery. We addressed this gap by generating multispectral images from Sentinel-1 to derive optical images, NDWI, and NDVI using a novel flow matching method. This, in the sequel, removes the cloud cover issue with respect to Sentinel-2 images through our model uniquely leveraging the resilience of Sentinel-1 in this context.  

To accomplish this task, we build upon the broader landscape of generative models. Now, these models typically follow one of two main paradigms. 
One such method is Generative Adversarial Networks (GANs) \cite{goodfellow2014generative}. 
In this setup, two models compete: one generates outputs similar to the target data, and the other tries to distinguish between real and generated data. However, GANs suffer from training instability among other issues \cite{wiatrak2019stabilizing}. Therefore, image-to-image GAN models, such as Pix2Pix \cite{henry2021pix2pix} and CycleGAN \cite{zhu2017unpaired} are not the top choice for image-to-image translation tasks.

Another category of methods aims to iteratively translate from one distribution to another, such as diffusion models \cite{sohl2015deep} and flow matching \cite{lipman2022flow}. Traditionally, these methods typically start from a noise distribution and gradually transform it into the target distribution. Notably, during training of these methods, diffusion goes from the target distribution to the noise distribution, and only in the reverse process of inference they do the opposite to get to the target distribution. On the other hand, flow matching moves directly from noise to target distribution in both training and inference. However, for deterministic translation tasks, such as image-to-image translation, starting from the input distribution is more appropriate. This choice avoids unnecessary meandering in the translation process and reduces the number of steps required.

Given that using the input distribution and target distribution as the two endpoints is a more efficient strategy, we now consider some of the common iterative models in image-to-image translation. The Brownian Bridge Diffusion Model (BBDM) \cite{li2023bbdm} learns a discrete sequence of steps to transition from the target distribution to the input distribution. During inference, the model reverses this process, subtracting the learned steps back from the input distribution to reach the target distribution. In contrast, Latent Flow Matching \cite{dao2023flow} learns a continuous transformation path from the input latent to the target latent distribution. Although these approaches differ in their training direction and granularity (discrete vs. continuous), fundamentally, both aim to model the transformation path between two distributions. One could argue that the training direction is a significant difference, but in essence, both methods are designed to learn this path. We further argue that Flow Matching is a superior approach due to its continuous nature, which allows it to be viewed as a superset of BBDM by covering more intermediate steps during training.

This brings us to the concept of the path itself—what exactly defines it? The answer lies in interpolation. In this context, interpolation refers to the scheduling mechanism that guides the model in determining how much to update its state at each step. This “update” quantifies the extent to which the model should translate towards the target distribution at a given time step. Importantly, the early steps in the process tend to be more error-prone. Therefore, assigning equal weight to all steps—as in linear scheduling—is suboptimal, particularly for tasks resembling optimal transport. To mitigate this, smaller updates should be applied in the early stages, motivating the adoption of non-linear scheduling strategies. Among various non-linear scheduling methods, we focus on two specific options: exponential and cosine scheduling. While the exponential and linear schedules 
were discussed in the original Flow Matching paper~\cite{lipman2022flow}, to the best of our knowledge, the cosine schedule has not been explored in prior literature of flow matching, including~\cite{lipman2022flow, lipman2024flow, dao2023flow}. We propose the cosine scheduling 
as a novel addition to the methods for flow matching in this context. 
Cosine scheduling, similar to exponential scheduling, assigns lower weights to the initial steps. However, the update magnitude  gradually increases toward the final steps. In other words, the effect of the predicted direction vector from initial distribution towards target distribution becomes more pronounced near the end, allowing for more significant changes as the model approaches the target distribution. As we will later see, each non-linear scheduling has its own benefits.  Additionally, we propose a novel modified multi-staged training procedure that specifically addresses the common issue of large errors during the initial stages of translation. Our method places greater emphasis on the early steps and aligns training more closely with the inference process. Furthermore, to stabilize and guide the translation, we condition the model on the initial distribution at every step.

\section{Results and Discussion}\label{sec:results}

\subsection{Methodical Overview}

\begin{figure*}[!htbp]
\begin{center}
    \subfigure[\label{fig:method_pipeline}]{
        \includegraphics[width=\textwidth]{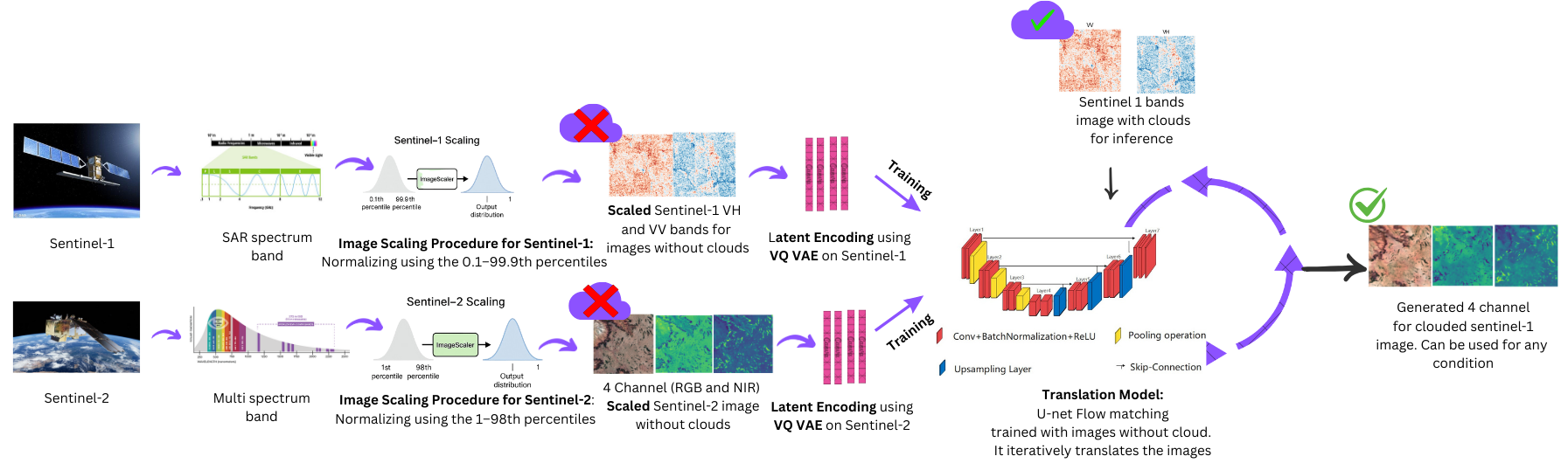}
    }
    \vspace{0.5cm}
    \subfigure[\label{fig:disaster_inference}]{
        \includegraphics[width=\textwidth]{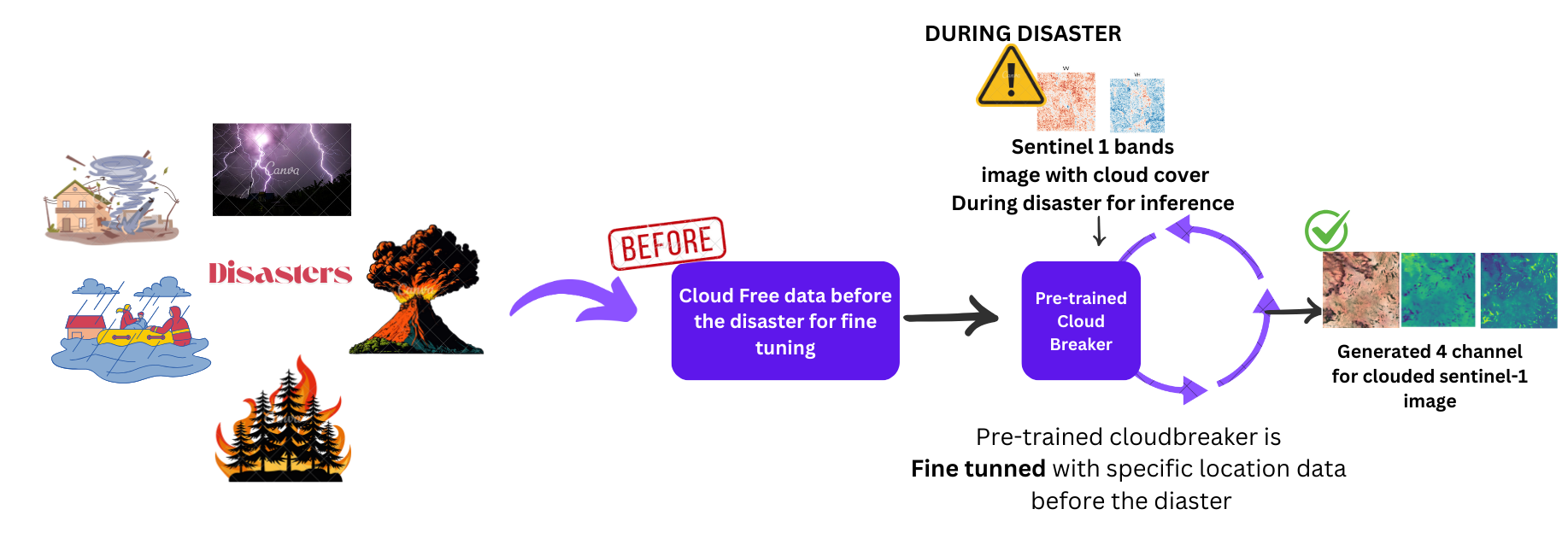}
    }
\end{center}
\caption{
    \textbf{a.} This subfigure illustrates the training procedure of our \textit{CloudBreaker} model. We use VQ-VAE latent space representations of cloud-free Sentinel-1 images as input and the corresponding Sentinel-2 latent representations as the ground truth for the U-Net model. After training with the flow matching method, CloudBreaker learns to directly generate Sentinel-2 latent representations from Sentinel-1 inputs. These are then decoded to reconstruct 4-channel images (RGB and Near-Infrared), which are used to compute optical (RGB), NDWI, and NDVI outputs. \textbf{b.} For practical deployment, we fine-tune the model using cloud-free data from the target region. Once fine-tuned, CloudBreaker can be used during real disaster events to generate cloud-free optical products from Sentinel-1 inputs.
}

\label{fig:methods}
\end{figure*}

In this study, we transform Sentinel-1 inputs into Sentinel-2 outputs via a learned mapping in a compact latent space. As shown in the Fig.~\ref{fig:method_pipeline} process is composed of several key stages.

\textbf{Image Scaling Procedure}

To standardize input distributions and improve training stability, all images were normalized per channel using a custom scaler named \texttt{ImageScaler}. Scaling was based on percentile thresholds \cite{MeVisLab_NormalizeImageByPercentileMapping,colom2013analysis} to reduce the influence of outliers. For Sentinel-1, the 0.1\textsuperscript{th} and 99.9\textsuperscript{th} percentiles were used; for Sentinel-2, we used the 1\textsuperscript{st} and 98\textsuperscript{th} percentiles. Each value was linearly scaled as follows (Eqn. \ref{eq:scaling}). No clipping was applied, preserving all values during training.
\begin{equation}
\label{eq:scaling}
\text{scaled} = \frac{\text{value} - \text{pmin}}{\text{pmax} - \text{pmin} + \varepsilon}
\end{equation}

\textbf{Latent Encoding.} Sentinel-1 and Sentinel-2 images were encoded using separate Vector-Quantized Variational Autoencoders (VQ-VAE)~\cite{van2017neural}, each reducing spatial resolution by a factor of 2 and projecting the inputs into a 16-channel latent space. This ensures that both 2-channel Sentinel-1 and 4-channel Sentinel-2 inputs are mapped to the same latent dimensionality, enabling computation of the difference vector $\Delta Z_s = Z_{S_2} - Z_{S_1}$, where $Z_{S_1}$ and $Z_{S_2}$ denote the latent representations of Sentinel-1 and Sentinel-2, respectively. This difference vector guides the model in learning the transformation from initial to target distribution.

\textbf{Translation Model.} The core model is a U-Net architecture \cite{ronneberger2015u} (\texttt{UNet2DModel}) that operates entirely within the latent space. Once the images were encoded into their latent forms, we trained the model following Algorithm~\ref{algo:cosine}. The training process included several notable techniques. First, we applied cosine-interpolated mixing: at each training step, we generated an intermediate latent code \(x_t\) as a weighted blend of \(Z_{S_1}\) and \(Z_{S_2}\), with the weight \(m_t\in[0,1]\) determined by a cosine schedule (Eq.~\ref{eq:cosine}). This approach allowed the model to learn smooth transitions from input to output with smaller updates initially and larger steps in the later
phases, improving generalization across intermediate representations.

To further enrich the training, we employed a multi-stage procedure that incorporated three types of sampling per batch, described as follows. In the first stage of our training, referred to as the \textit{continuous} mode, \(m_t\) was uniformly sampled from the interval \([0,1]\) to generate a random \(x_t\) from the continuous path for each example. In the second stage, referred to as the \textit{discrete} mode, an integer \(t \in \{0, \dots, N-1\}\) was chosen and we set \(m_t = t/N\) to ensure coverage of fixed points along the interpolation path. Lastly, in the \textit{boundary focus} mode, we always included \(m_t = 0\) (i.e., pure \(Z_{S_1}\)) to strengthen the model’s learning of the most difficult transformation.


\textbf{Post-processing \& Indices.} After inference, we used the steps described in Algorithm~\ref{algo:inference} to decode the predicted latent representation \(\hat Z_{S_2}\) into a reconstructed Sentinel-2 image. We then separated the RGB and NIR channels to compute two indices, namely, the Normalized Difference Water Index (NDWI, Eq.~\ref{eq:NDWI}) and the Normalized Difference Vegetation Index (NDVI, Eq.~\ref{eq:NDVI}). Although we have used a global coverage dataset (see the ``\textit{Datasets}" Section in the \textit{Supplementary Material}, it does not cover every single environment in the world. So, for practical applications on a particular location, we fine-tune the model with some limited cloud-free data of those locations, as shown in Fig.~\ref{fig:disaster_inference}.

\begin{algorithm}
\caption{Training with Cosine Interpolation from Latent Space of Sentinel-1 to Latent Space of Sentinel-2}
\label{algo:cosine}
\begin{algorithmic}[1]
\State \textbf{Input:} Number of epochs $E$, number of interpolation steps $N$, random noise factor $r$, data loaders $\texttt{train\_latent\_loader}$ and $\texttt{val\_latent\_loader}$ (containing latent embeddings of Sentinel-1 and Sentinel-2), model $\mathcal{M}$, optimizer, scheduler
\State \textbf{Initialize:} Time step size $\Delta t \gets \frac{1}{N}$

\For{each epoch $e = 1$ to $E$}
    \State Set model to training mode: $\mathcal{M} \gets \texttt{train mode}$
    \State Initialize total training loss: $L_{\text{total}} \gets 0$
    
    \For{each batch $(\mathbf{z}_{\text{S1}}, \mathbf{z}_{\text{S2}})$ in \texttt{train\_latent\_loader}}
        \Comment{--- \textbf{Training on continuous steps} ---}
        \State Sample random timestep: $m_t \sim \mathcal{U}(0, 1)$
        \State Compute interpolated latent: 
        \[
        \mathbf{x}_t \gets \frac{1}{2}(1 - \cos(\pi m_t)) \cdot \mathbf{z}_{\text{S2}} + \left(1 - \frac{1}{2}(1 - \cos(\pi m_t))\right) \cdot \mathbf{z}_{\text{S1}}
        \]
        \State Predict: $\hat{\mathbf{y}} \gets \mathcal{M}([\mathbf{x}_t, \mathbf{z}_{\text{S1}}], m_t)$
        \State Set target: $\mathbf{y} \gets \mathbf{z}_{\text{S2}} - \mathbf{z}_{\text{S1}}$
        \State Compute loss: $\mathcal{L} \gets \|\mathbf{y} - \hat{\mathbf{y}}\|^2$
        \State Update model parameters using $\mathcal{L}$
        \State $L_{\text{total}} \gets L_{\text{total}} + \mathcal{L}$

        \Comment{--- \textbf{Training on discrete time steps} ---}
        \State Sample $t \sim \{0, \dots, N - 1\}, \quad m_t \gets \frac{t}{N}$
        \State Repeat lines 8--13 with the new $m_t$

        \Comment{--- \textbf{Training at initial step} ($t = 0$) ---}
        \State Predict: $\hat{\mathbf{y}} \gets \mathcal{M}([\mathbf{z}_{\text{S1}}, \mathbf{z}_{\text{S1}}], m_t = 0)$
        \State Compute loss: $\mathcal{L} \gets \|\mathbf{z}_{\text{S2}} - \mathbf{z}_{\text{S1}} - \hat{\mathbf{y}}\|^2$
        \State Update model parameters
        \State $L_{\text{total}} \gets L_{\text{total}} + \mathcal{L}$
    \EndFor

    \State Compute average training loss: $\bar{L}_{\text{train}} \gets \frac{L_{\text{total}}}{\text{num\_batches}}$
    \State Set model to evaluation mode: $\mathcal{M} \gets \texttt{eval mode}$, initialize validation loss: $L_{\text{val}} \gets 0$

    \For{each batch $(\mathbf{z}_{\text{S1}}, \mathbf{z}_{\text{S2}})$ in \texttt{val\_latent\_loader}}
        \State Sample $t \sim \{0, \dots, N - 1\}, \quad m_t \gets \frac{t}{N}$
        \State Interpolate:
        \[
        \mathbf{x}_t \gets \frac{1}{2}(1 - \cos(\pi m_t)) \cdot \mathbf{z}_{\text{S2}} + \left(1 - \frac{1}{2}(1 - \cos(\pi m_t))\right) \cdot \mathbf{z}_{\text{S1}}
        \]
        \State Predict: $\hat{\mathbf{y}} \gets \mathcal{M}([\mathbf{x}_t, \mathbf{z}_{\text{S1}}], m_t)$
        \State Compute validation loss:
        \[
        \mathcal{L} \gets \|\mathbf{z}_{\text{S2}} - \mathbf{z}_{\text{S1}}  - \hat{\mathbf{y}}\|^2
        \]
        \State $L_{\text{val}} \gets L_{\text{val}} + \mathcal{L}$
    \EndFor

    \State Compute average validation loss: $\bar{L}_{\text{val}} \gets \frac{L_{\text{val}}}{\text{num\_batches}}$
    \State Step the scheduler using $\bar{L}_{\text{val}}$
    
    \If{epoch $e$ is a checkpoint interval}
        \State Save model checkpoint
    \EndIf
\EndFor
\end{algorithmic}
\end{algorithm}

\begin{algorithm}
\caption{Inference: Latent-Space Translation with Cosine Schedule}
\label{algo:inference}
\begin{algorithmic}[1]
\State \textbf{Input:} pretrained model $\mathcal{M}$, Sentinel-2 VQ-VAE decoder $\mathrm{Dec}$, test loader $\mathit{test\_latent\_loader}$, steps $T$
\State \textbf{Output:} reconstructed RGB/NIR images, NDVI, NDWI

\For{each batch $(\mathbf{z}_{\mathrm{S1}}, \mathbf{z}_{\mathrm{S2}})$ in $\mathit{test\_latent\_loader}$}
  \State $\mathbf{x} \gets \mathbf{z}_{\mathrm{S1}}$ \Comment{initialize at S1 latent}
  \State Compute cosine schedule: 
    $s_i = \tfrac12\bigl(1 - \cos(\tfrac{\pi\,i}{T-1})\bigr),\; i = 0,\dots,T-1$
  \For{$i = 0$ to $T-2$}
    \State $m \gets s_i$
    \State $\delta \gets s_{i+1} - s_i$
    \State $\Delta \mathbf{z} \gets \mathcal{M}\bigl([\mathbf{x},\,\mathbf{z}_{\mathrm{S1}}],\,m\bigr)$
    \State $\mathbf{x} \gets \mathbf{x} + \Delta \mathbf{z}.\mathrm{sample} \times \delta$
  \EndFor

  \State \textbf{Decode:} $\hat{\mathbf{X}} \gets \mathrm{Dec}(\mathbf{x})$ \Comment{shape: (B,4,H,W)}
  \State Split channels: 
    $\hat{\mathbf{X}}_{\mathrm{RGB}} = \hat{\mathbf{X}}[:,0:3,:,:]$, 
    $\hat{\mathbf{X}}_{\mathrm{NIR}} = \hat{\mathbf{X}}[:,3:4,:,:]$
  \State Compute indices ($\epsilon=10^{-7}$): 
    $\mathrm{NDVI} = (\hat{\mathbf{X}}_{\mathrm{NIR}} - \hat{\mathbf{X}}_{\mathrm{RGB}}[:,0:1,:,:]) / (\hat{\mathbf{X}}_{\mathrm{NIR}} + \hat{\mathbf{X}}_{\mathrm{RGB}}[:,0:1,:,:] + \epsilon)$, 
    $\mathrm{NDWI} = (\hat{\mathbf{X}}_{\mathrm{RGB}}[:,1:2,:,:] - \hat{\mathbf{X}}_{\mathrm{NIR}}) / (\hat{\mathbf{X}}_{\mathrm{RGB}}[:,1:2,:,:] + \hat{\mathbf{X}}_{\mathrm{NIR}} + \epsilon)$

  \State Save or return $\hat{\mathbf{X}}_{\mathrm{RGB}},\,\hat{\mathbf{X}}_{\mathrm{NIR}},\,\mathrm{NDVI},\,\mathrm{NDWI}$
\EndFor
\end{algorithmic}
\end{algorithm}

\begin{equation}
\label{eq:linear}
x_t = (1 - m_t) \cdot s_1 + m_t \cdot s_2, \quad \text{where } m_t \in [0, 1]
\end{equation}

\begin{equation}
\label{eq:expo}
x_t = \left(1 - w_t\right) \cdot s_1 + w_t \cdot s_2, \quad \text{where } w_t = \frac{e^{k (m_t - 1)} - \min}{\max - \min + \varepsilon}, \quad m_t \in [0, 1]
\end{equation}

\begin{equation}
\label{eq:cosine}
x_t = \left(1 - \frac{1}{2} \left(1 - \cos(\pi \cdot m_t)\right)\right) \cdot s_1 + \frac{1}{2} \left(1 - \cos(\pi \cdot m_t)\right) \cdot s_2, \quad \text{where } m_t \in [0, 1]
\end{equation}

\begin{equation}
\label{eq:NDWI}
\text{NDWI} = \frac{\text{Green} - \text{NIR}}{\text{Green} + \text{NIR}}
\end{equation}
\begin{equation}
\label{eq:NDVI}
\text{NDVI} = \frac{\text{NIR} - \text{Red}}{\text{NIR} + \text{Red}}
\end{equation}

\begin{figure*}[t]

\begin{center}
     \includegraphics[width=\textwidth]{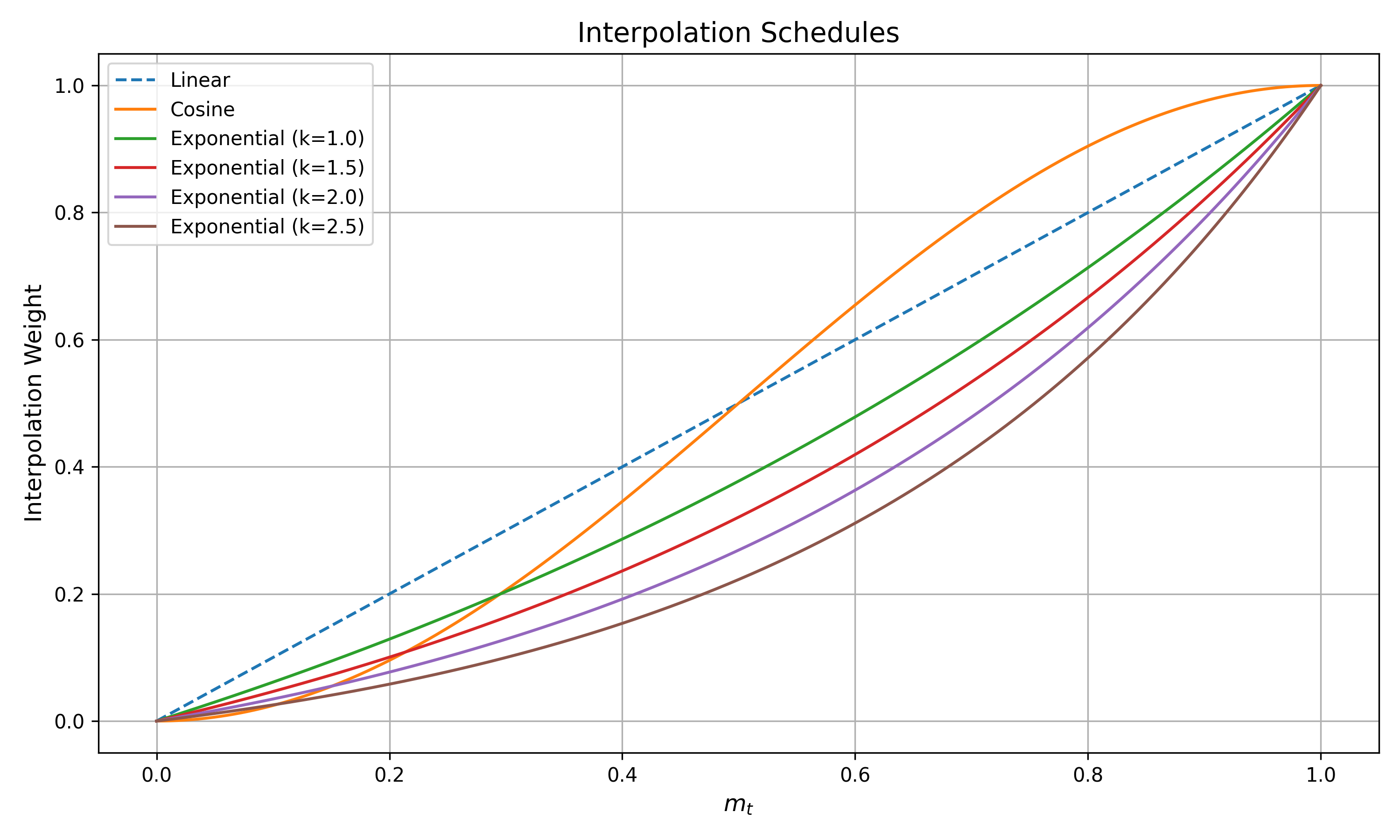}

\end{center}
   \caption{Different Scheme of interpolation or scheduling}
    \label{fig:scheduling_scheme}
\end{figure*}

\subsection{Latent Space results}
Recall that using two VQ-VAEs, 
we successfully converted Sentinel-1 and Sentinel-2 images into a latent space. The reconstruction performance from the latent space was evaluated on the decoded outputs. For decoded Sentinel-2, we achieved a mean squared error (MSE) of 0.0006 and a $R^2$ score of 0.985. For decoded Sentinel-1, the model achieved an MSE of 0.00054 and a $R^2$ score of 0.97. These results indicate that the latent encoding effectively preserved the essential information from the original images, allowing for high-precision reconstruction.

\subsection{Translation Results}
We evaluated our model in two stages: first, in the latent space, where the model learns the mapping (translation) between Sentinel-1 and Sentinel-2 representations; and second, in the decoded image space, where the 4-channel (RGB + NIR) Sentinel-2 output is reconstructed. From the reconstructed images, we further compute and evaluate domain-specific remote sensing indices, such as NDVI and NDWI (which will be shown in the following section).

As shown in Table~\ref{tab:scheduler_metrics}, we compare GAN-based methods, such as Pix2Pix and CycleGAN, diffusion-based methods, such as BBDM, and various configurations of Flow Matching (FM). The evaluation metrics include Fréchet Inception Distance (FID) \cite{dowson1982frechet,heusel2017gans}, Structural Similarity Index (SSIM) \cite{nilsson2020understanding}, Learned Perceptual Image Patch Similarity (LPIPS) \cite{zhang2018unreasonable}, as well as MSE and $R^2$ (see Section~\ref{subsec:metric_used} in the Supplementary Materials). 

From the results, we observe that FM methods outperform the other approaches by a substantial margin. GAN-based methods generally perform the worst across most metrics, while BBDM comes closer to FM in performance, but still lags behind, particularly in RGB FID (second worst) and RGB LPIPS (worst). 

\begin{figure}[htbp]
    \centering
    \includegraphics[width=\textwidth]{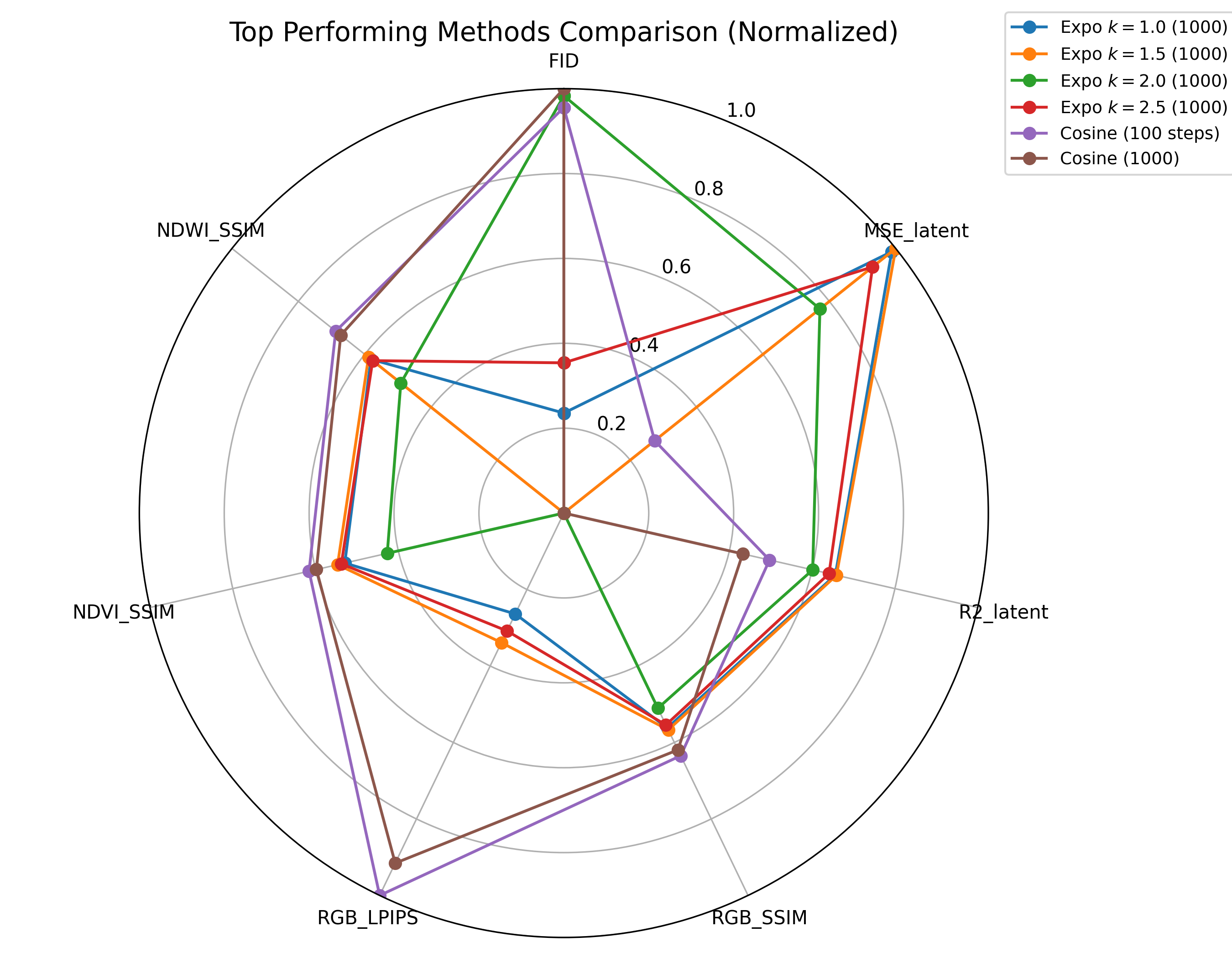}
    \caption{
    Radar plot comparison of top-performing methods on the test dataset, normalized across seven evaluation metrics: FID, MSE\_latent, R$^2$\_latent, RGB SSIM, RGB LPIPS, NDVI SSIM, and NDWI SSIM. The Cosine (1000) scheduler achieves the best perceptual quality with the lowest FID (0.6300) and RGB LPIPS (0.3433), outperforming Expo $k=2.5$ (1000) by 68.9\% in FID and 4.5\% in LPIPS. While Expo $k=1.5$ (1000) yields the highest R$^2$\_latent (0.6588) and strong NDWI SSIM (0.5886), the Cosine scheduler offers superior perceptual realism. These results indicate a trade-off between perceptual and structural metrics, with the Cosine scheduler emerging as the best choice when visual quality is prioritized.
    }
    \label{fig:top_methods_comparison}
\end{figure}

Additionally, we comprehensively evaluated the FM model using different schedulers defined in Eq.~\ref{eq:linear}, Eq.~\ref{eq:expo}, and Eq.~\ref{eq:cosine}, under two inference step settings (100 and 1000), as summarized in Table~\ref{tab:scheduler_metrics}. From the radar plot~\cite{saary2008radar} comparing the top-performing models across multiple metrics in Fig.~\ref{fig:top_methods_comparison}, we observe that the cosine scheduler (1000 steps) achieves the best perceptual quality, with the lowest FID of \textbf{0.6481}, outperforming all other models by more than 50\%. Compared to the next best model, the exponential scheduler (Expo $k=2$, FID = 0.6840), the cosine scheduler (1000 steps) shows an improvement of approximately \textbf{5.25\%} in FID. While exponential schedulers, such as Expo $k=1.5$ demonstrate a more balanced performance across certain evaluation metrics— achieving the highest $R^2$ score and a low latent-space MSE— the cosine scheduler excels in key structural quality measures, including SSIM of NDVI, SSIM of NDWI, and LPIPS of RGB. This consistent superiority in structural fidelity makes the cosine scheduler the most suitable choice for applications where perceptual realism and visual quality are of primary importance, especially in scenarios involving human interpretation. Additionally, from Table~\ref{tab:scheduler_metrics}, we also observe that the performance of exponential scheduling is highly sensitive to the parameter $k$, which controls the steepness of the scheduling curve. Based on metric-specific priorities, users can select models accordingly. Since inference time is a critical consideration for practical applications, we prioritize models that maintain strong performance with fewer inference steps (e.g., 100 steps), as they offer faster generation while still producing realistic images as indicated by lower FID scores. Therefore, our main discussion focuses on cosine interpolation with 100 steps. 


\subsubsection{Translated Sentinel-2 Latent Space Evaluation}
To assess how effectively the model learns the transformation in the latent space, we compute the Mean Squared Error (MSE) and the coefficient of determination (\(R^2\)) between the predicted latent representation from Sentinel-1 and the ground truth latent embedding of Sentinel-2. The model achieves an MSE of \textbf{0.003814} and a \(R^2\) score of \textbf{0.4969}, indicating a reasonable alignment between the predicted and target latent distributions. However, since the $R^2$ score overlooks nonlinear, perceptual, and structural factors that are critical for evaluating generated images, it is not the preferred metric for such tasks. Please refer to Section~\ref{subsec:metric_used} in the Supplementary for more details.

\subsubsection{Reconstructed Image Evaluation}
After decoding the latent representation, we obtain the full-resolution 4-channel Sentinel-2 image, consisting of RGB and NIR bands. We evaluated the RGB subset using both perceptual and pixel-wise metrics to comprehensively assess reconstruction quality. The reconstructed RGB images achieved a Fréchet Inception Distance (FID) of \textbf{0.7432}, which is significantly low and indicates high perceptual similarity to real Sentinel-2 images—suggesting that the generative model effectively captured the high-level visual features.

In terms of pixel-wise evaluation, we observe a mean squared error (MSE) of \textbf{0.0198}, which reflects a low average error between reconstructed and ground truth pixel values. The peak signal-to-noise ratio (PSNR) is \textbf{17.29 dB}, indicating an acceptable level of reconstruction accuracy, especially given the complexity of natural scenes in satellite imagery. The SSIM (range 0 to 1 and the higher the better) is \textbf{0.6346}, which shows that essential structural patterns and textures are reasonably well preserved. Additionally, the LPIPS (range 0 to 1 and the lower the better) score of \textbf{0.2719} suggests good perceptual similarity at the feature level, as seen by a neural network. Together, these results demonstrate a very high visual fidelity with moderate structural consistency, confirming that the model was able to effectively reconstruct complex natural features.


\subsubsection{Remote Sensing Index Evaluation}
From the reconstructed 4-channel Sentinel-2 image, we computed NDVI and NDWI.  
These indices are derived from the NIR and visible bands, and help assess vegetation and water body coverage, respectively. Our reconstructed NDVI and NDWI achieved SSIM of \textbf{0.6156} and \textbf{0.6874} respectively, indicating a good level of structural similarity to the corresponding ground truth indices. SSIM ranges from 0 to 1, where higher values signify closer structural resemblance. For comparison, we evaluated the output of Ahmed et al.'s model \cite{sakib2025light}, which only generates NDWI, and obtained an NDWI SSIM of 0.423. This indicates that our model outperforms theirs in preserving the structural features of water-related spectral content in the reconstructed images. The improved SSIM demonstrates better spatial fidelity in our generated NDWI, highlighting the effectiveness of our approach in capturing fine-grained Earth surface characteristics.

\subsection{Removing Clouds in Test Set} 
\begin{figure}[htbp]
    \centering
    \subfigure[]{\includegraphics[width=0.3\textwidth]{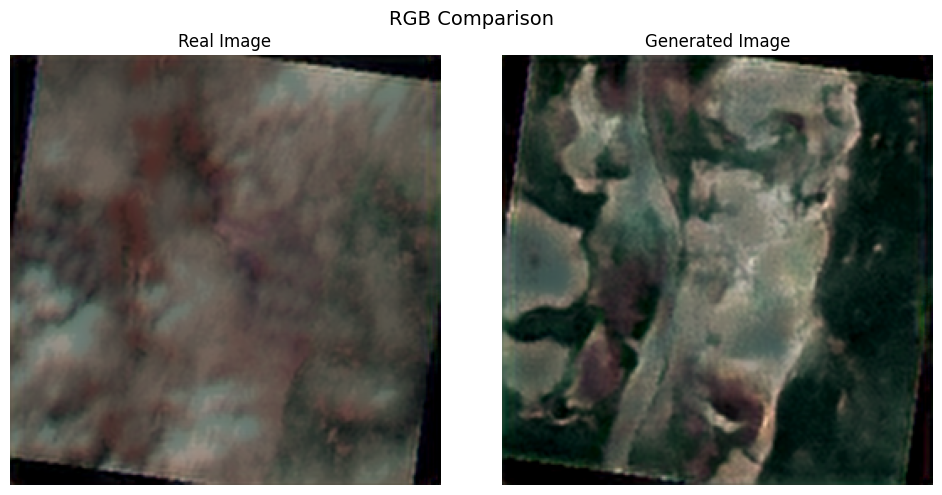}}
    \subfigure[]{\includegraphics[width=0.3\textwidth]{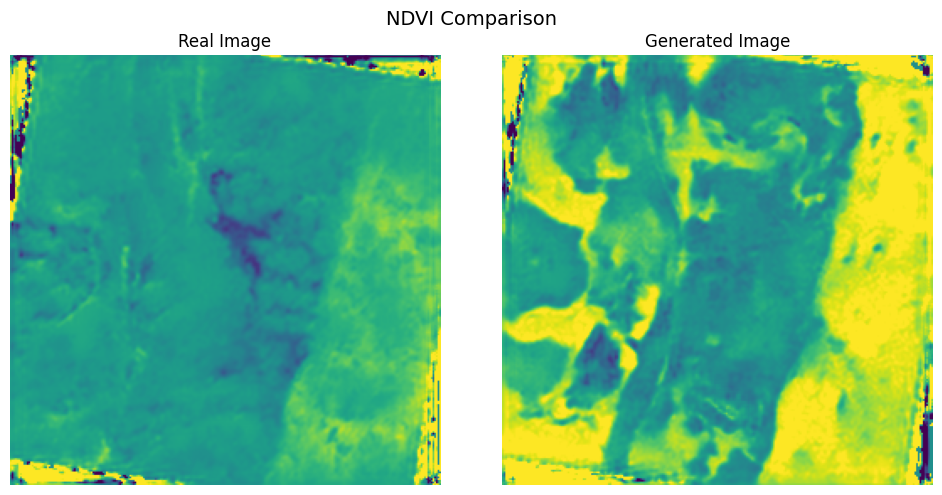}}
    \subfigure[]{\includegraphics[width=0.3\textwidth]{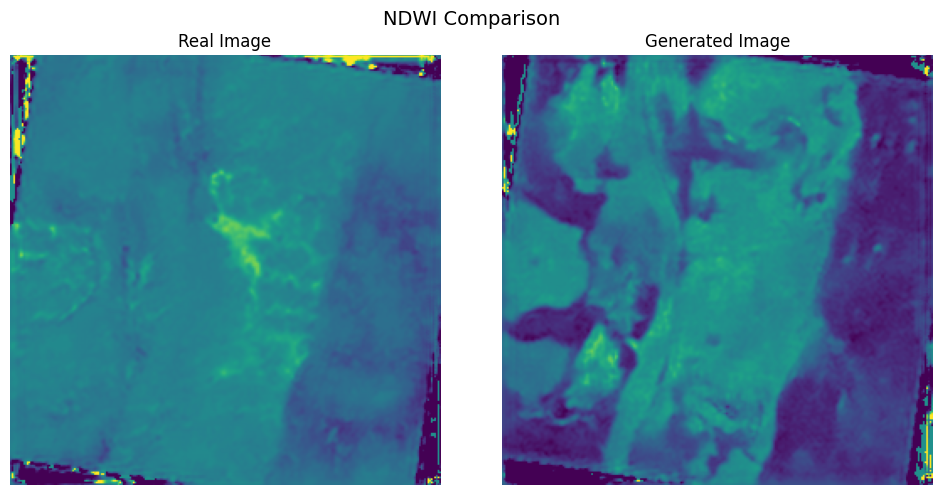}} \\
    \subfigure[]{\includegraphics[width=0.3\textwidth]{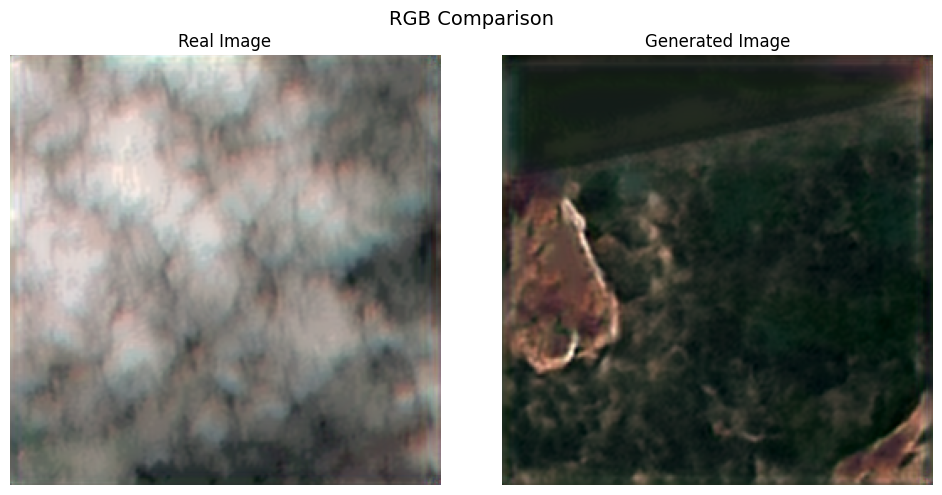}}
    \subfigure[]{\includegraphics[width=0.3\textwidth]{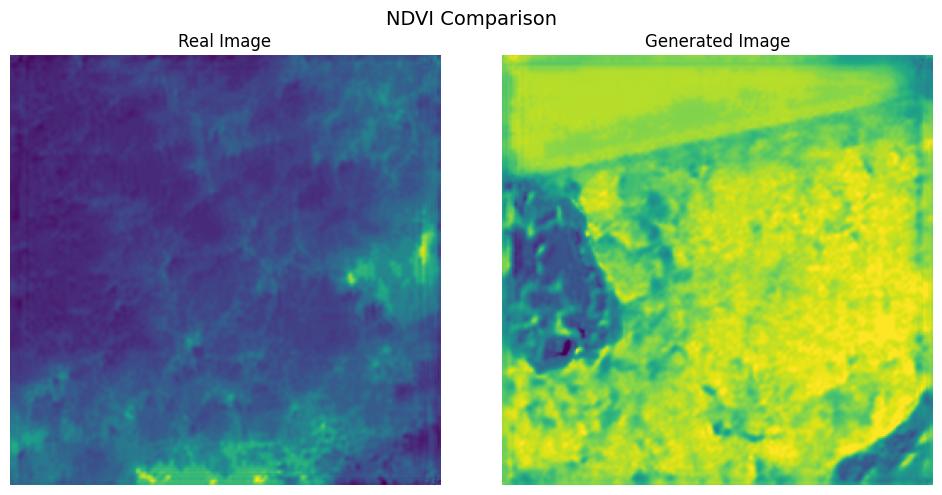}}
    \subfigure[]{\includegraphics[width=0.3\textwidth]{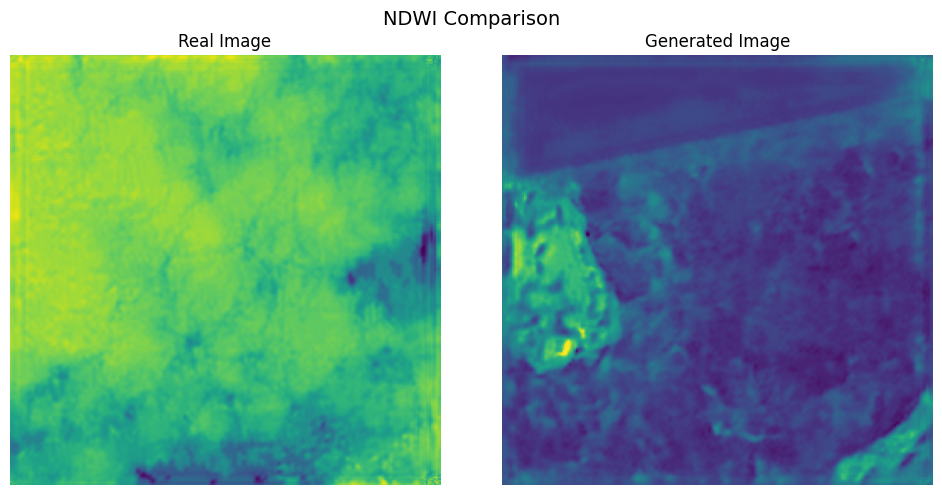}} \\
    \subfigure[]{\includegraphics[width=0.3\textwidth]{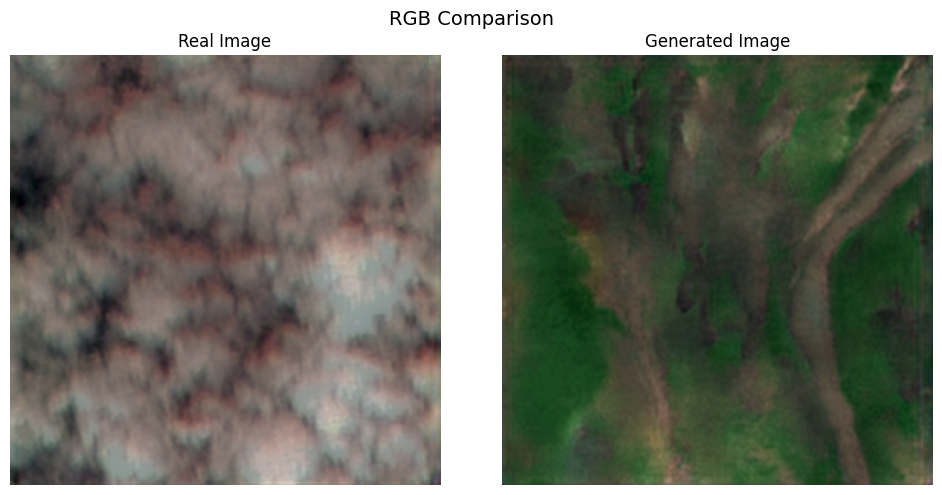}}
    \subfigure[]{\includegraphics[width=0.3\textwidth]{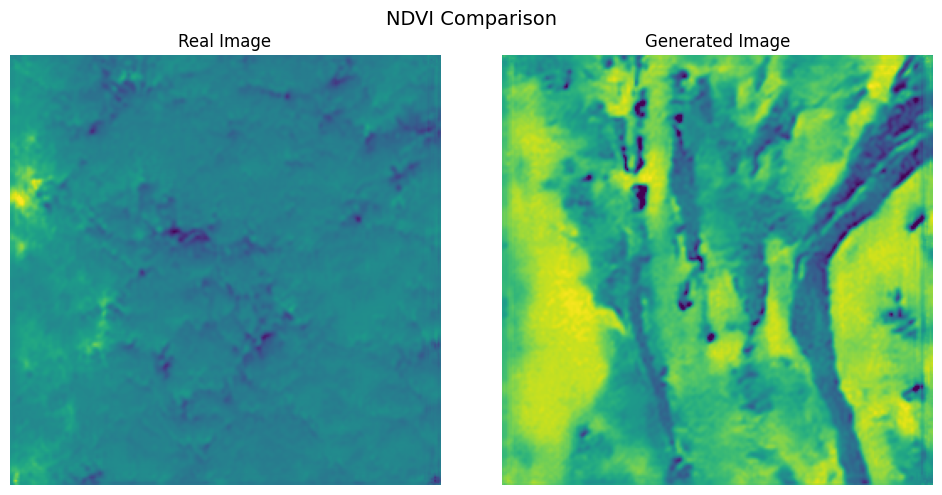}}
    \subfigure[]{\includegraphics[width=0.3\textwidth]{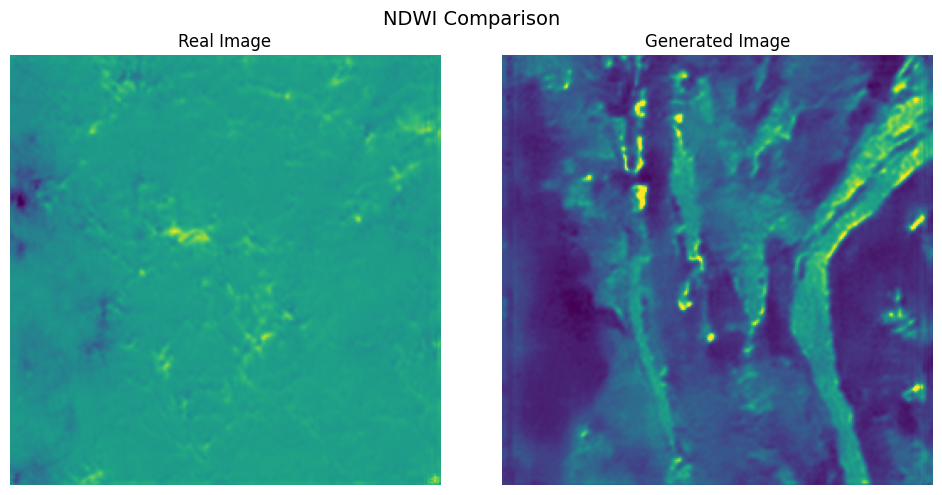}} \\

    \caption{\textbf{a, d, g:} Comparison between cloudy real (left) and generated (right) cloud free RGB images. \textbf{b, e, h:} Comparison between cloudy real (left) and generated (right) cloud free NDVI images. \textbf{c, f, i:} Comparison between cloudy real (left) and generated (right) cloud free NDWI images.}

    \label{fig:cloudy_images}
\end{figure}

As we can see from Fig.~\ref{fig:cloudy_images}, the use of CloudBreaker has successfully removed the clouds from the test satellite images. The first column shows the optical (RGB) images, while the second and third columns present the corresponding NDVI and NDWI representations for three test samples. Each subfigure presents a pair of images— the left side showing the original cloud-covered image and the right side showing the corresponding image after cloud removal.

In the RGB images (Fig.~\ref{fig:cloudy_images}.a, Fig.~\ref{fig:cloudy_images}.d, and Fig.~\ref{fig:cloudy_images}.g), we can now clearly observe underlying surface features and land cover structures that were previously obscured due to cloud cover. This clarity is crucial for accurate visual interpretation and analysis of spatial features.

Furthermore, the presence of clouds in the original images negatively affected the quality of the derived vegetation and water indices. The NDVI (shown in Fig.~\ref{fig:cloudy_images}.b, Fig.~\ref{fig:cloudy_images}.e, and Fig.~\ref{fig:cloudy_images}.h) and NDWI (shown in Fig.~\ref{fig:cloudy_images}.c, Fig.~\ref{fig:cloudy_images}.f, and Fig.~\ref{fig:cloudy_images}.i) were distorted due to cloud interference, resulting in incomplete or misleading information. However, after removing the clouds, these indices were properly recovered, revealing the true vegetation health and water distribution in the region. This demonstrates the importance of cloud removal as a preprocessing step for reliable satellite image analysis in environmental monitoring and remote sensing applications.
\subsection{Real-life Case Study}
We have also tested the model for various real-life disastrous events. As it can be observed that Fig.~\ref{fig:disaster_RGB_comparison}, together with Fig.~\ref{fig:disaster_NDVI_comparison} and Fig.~\ref{fig:disaster_NDWI_comparison} in the supplementary materials, illustrate the optical (RGB), NDVI, and NDWI representations, respectively, of the various disaster events analyzed below. Each figure is organized into five rows corresponding to five different disaster events. Each row contains three columns that represent the satellite imagery before (first column), during (second column), and after (third column) the event. Within each panel (cell), two images are shown side by side: the left image is the original satellite observation (which may include cloud cover, especially in the “during” images, as is common during disasters), and the right image is the corresponding output generated by our model. This structure provides a consistent visual comparison across the different data modalities, highlighting both the impact of each disaster and the effectiveness of the model in reconstructing clearer views under challenging conditions. Below, we present the results of our model in these real-world scenarios.

\subsubsection{Amazon Fire}
Our first real-life example is the Amazon fire that occurred in July 2023. This devastating fire scorched large areas of rainforest, destroyed countless trees, and threatened vulnerable wildlife. It led to severe biodiversity loss, worsened air quality, and released massive carbon emissions. To analyze this event, we used Sentinel-1 imagery from before, during, and after the disaster, removing clouds for better clarity. The first row of Fig.~\ref{fig:disaster_RGB_comparison} (main text) and Fig.~\ref{fig:disaster_NDVI_comparison} and Fig.~\ref{fig:disaster_NDWI_comparison} (supplementary materials) illustrates the Amazon fire event in optical (RGB), NDVI, and NDWI representations, respectively. Each figure includes the “before” stage: Fig.~\ref{fig:disaster_RGB_comparison}.a (main text) and Fig.~\ref{fig:disaster_NDVI_comparison}.a and Fig.~\ref{fig:disaster_NDWI_comparison}.a (supplementary materials), where the left column shows the original satellite observation and the right column shows the corresponding output generated by our model. These images clearly depict the intact forest area prior to the fire, with NDVI and NDWI effectively highlighting vegetation health and surface water content, respectively. The b panel — Fig.~\ref{fig:disaster_RGB_comparison}.b (main text) and Fig.~\ref{fig:disaster_NDVI_comparison}.b and Fig.~\ref{fig:disaster_NDWI_comparison}.b (supplementary materials) — presents the corresponding images during the fire, highlighting the impact on vegetation and water content. The NDVI and NDWI values visibly decline, reflecting the degradation of forest cover and moisture content. Finally, the c panel — Fig.~\ref{fig:disaster_RGB_comparison}.c (main text) and Fig.~\ref{fig:disaster_NDVI_comparison}.c and Fig.~\ref{fig:disaster_NDWI_comparison}.c (supplementary materials) — shows the area after the fire. A significant reduction in vegetation and water content can be observed in our generated image in the right of those respective figures. These could not be seen in the original images in the left due to being obscured by cloud cover. These visualizations underscore the utility of our model in monitoring and analyzing the progression and aftermath of environmental disasters.

\subsubsection{Hurricane Harvey}
We now turn to our second disaster, Hurricane Harvey \cite{hurricane_harvey_wikipedia}, that occurred in the USA in 2017. Hurricane Harvey struck Texas and Louisiana in August 2017, causing catastrophic flooding that displaced over 30,000 people and resulted in more than \$125 billion in damages, making it the costliest natural disaster in Texas history . With peak rainfall exceeding 60 inches, it also became the wettest tropical cyclone ever recorded in the United States. The results of applying our model to remove clouds from satellite imagery captured before, during, and after this event are presented below. The second row of Fig.~\ref{fig:disaster_RGB_comparison} (main text) and Fig.~\ref{fig:disaster_NDVI_comparison} and Fig.~\ref{fig:disaster_NDWI_comparison} (supplementary materials) illustrates the Hurricane Harvey in optical (RGB), NDVI, and NDWI representations, respectively. Fig.~\ref{fig:disaster_RGB_comparison}.d (main text) and Fig.~\ref{fig:disaster_NDVI_comparison}.d and Fig.~\ref{fig:disaster_NDWI_comparison}.d (supplementary materials) show the optical (RGB), NDVI, and NDWI images of the area before the hurricane. These images clearly delineate the pre-disaster landscape, with NDVI and NDWI effectively capturing healthy vegetation and normal water content. The Fig.~\ref{fig:disaster_RGB_comparison}.e (main text) and Fig.~\ref{fig:disaster_NDVI_comparison}.e and Fig.~\ref{fig:disaster_NDWI_comparison}.e (supplementary materials) presents the corresponding images during the hurricane, highlighting the disruption to vegetation.  NDVI and NDWI values decrease noticeably, reflecting vegetation damage. A noticeable change in land cover can be observed in the generated images on the right, which are not clearly visible in the original images on the left due to cloud cover. Finally, Fig.~\ref{fig:disaster_RGB_comparison}.f (main text) and Fig.~\ref{fig:disaster_NDVI_comparison}.f and Fig.~\ref{fig:disaster_NDWI_comparison}.f (supplementary materials) show the area after the hurricane. 


\subsubsection{Nepal Flood}
Thirdly, we move to the Nepal flooding that happened in 2024 \cite{NepalFloods2024}. Torrential monsoon rains in July, August, and especially late September 2024 caused devastating floods and mudslides across Nepal, severely impacting infrastructure, homes, and agriculture. The floods resulted in over 224 deaths, thousands displaced, widespread damage to bridges, highways, hydropower stations, and disrupted essential services including power and telecommunications. The results from our model for this event are presented in the third row of Fig.~\ref{fig:disaster_RGB_comparison} (main text) and Fig.~\ref{fig:disaster_NDVI_comparison} and Fig.~\ref{fig:disaster_NDWI_comparison} (supplementary materials). Fig.~\ref{fig:disaster_RGB_comparison}.g (main text) and Fig.~\ref{fig:disaster_NDVI_comparison}.g and Fig.~\ref{fig:disaster_NDWI_comparison}.g (supplementary materials) present the optical (RGB), NDVI, and NDWI images of the area before the flood. These images clearly delineate the pre-flood landscape, with NDVI and NDWI effectively capturing healthy vegetation and normal water content.  Fig.~\ref{fig:disaster_RGB_comparison}.h (main text) and Fig.~\ref{fig:disaster_NDVI_comparison}.h and Fig.~\ref{fig:disaster_NDWI_comparison}.h (supplementary materials) present the corresponding images during the flood, highlighting the inundation and disruption to vegetation. The NDVI and NDWI values decline noticeably, reflecting the loss of vegetation and increased water coverage. A significant change in land cover and persistent water-logged areas can be observed in the generated images on the right. These details are not visible in the original images on the left due to cloud cover. Finally, Fig.~\ref{fig:disaster_RGB_comparison}.i (main text) and Fig.~\ref{fig:disaster_NDVI_comparison}.i and Fig.~\ref{fig:disaster_NDWI_comparison}.i (supplementary materials) show the area after the flood. 


\subsubsection{Cyclone Remal}
\begin{figure}[htbp]
    \centering
    \subfigure[]{\includegraphics[width=0.3\textwidth]{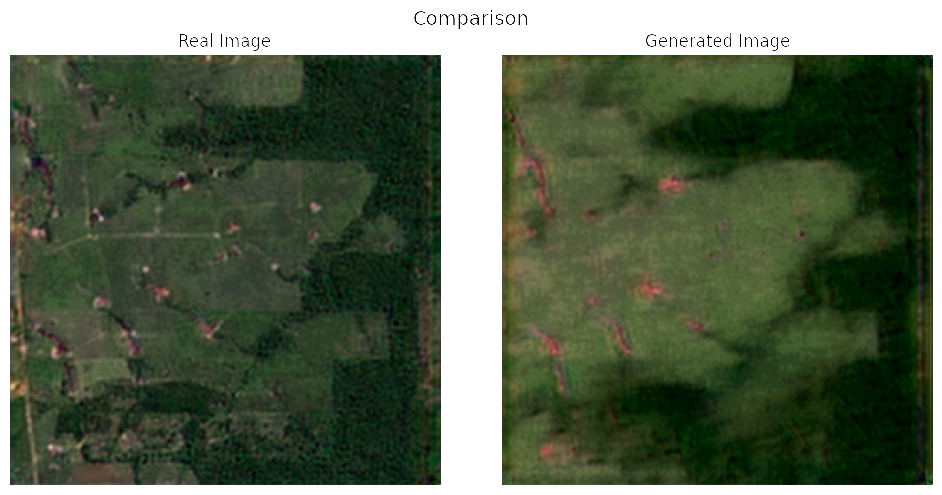}}
    \subfigure[]{\includegraphics[width=0.3\textwidth]{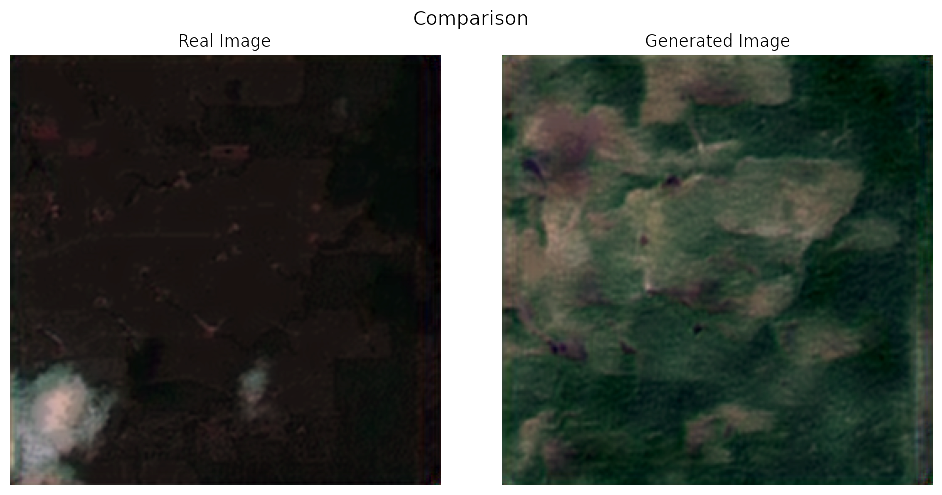}}
    \subfigure[]{\includegraphics[width=0.3\textwidth]{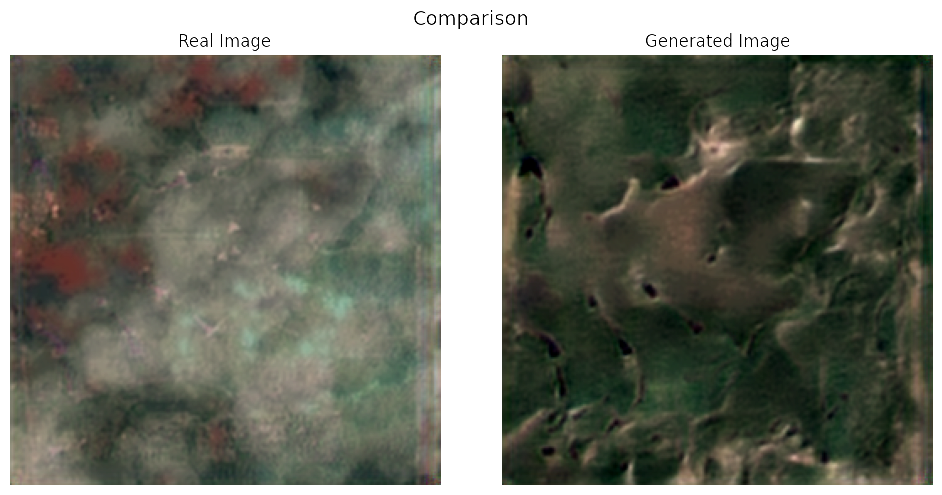}} \\[0.5cm]
    \subfigure[]{\includegraphics[width=0.3\textwidth]{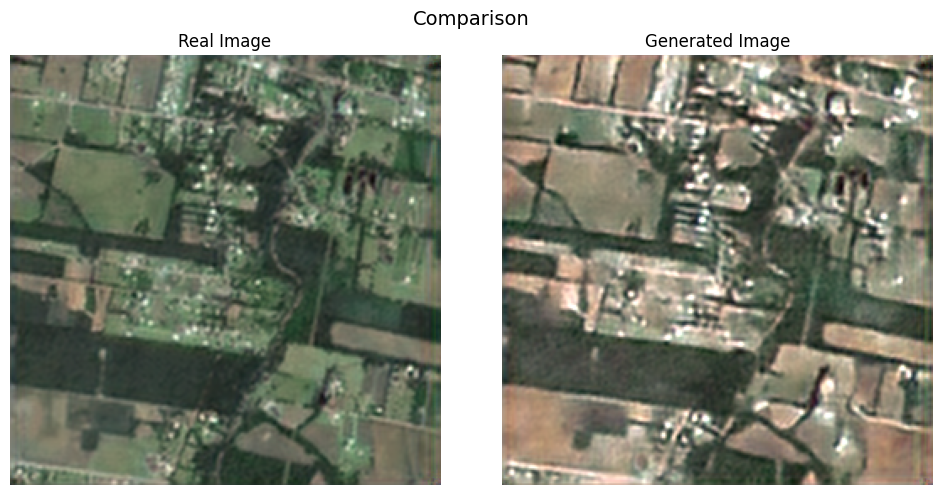}}
    \subfigure[]{\includegraphics[width=0.3\textwidth]{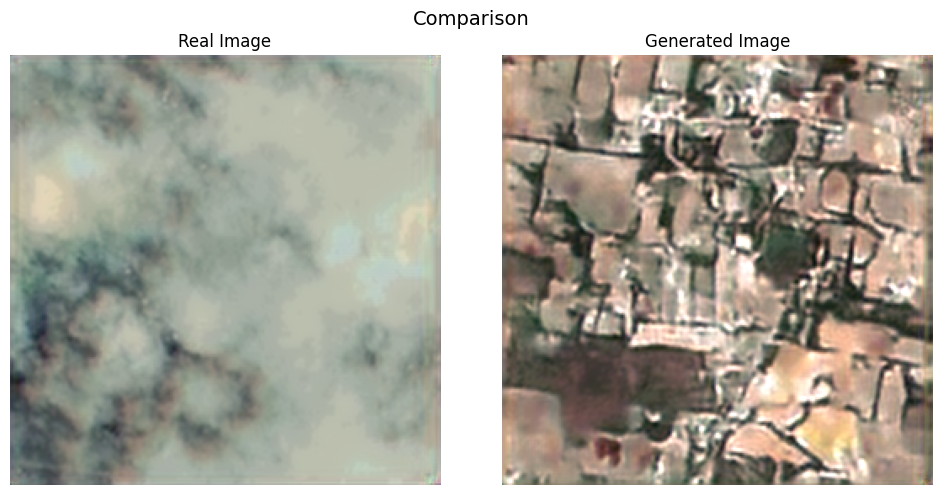}}
    \subfigure[]{\includegraphics[width=0.3\textwidth]{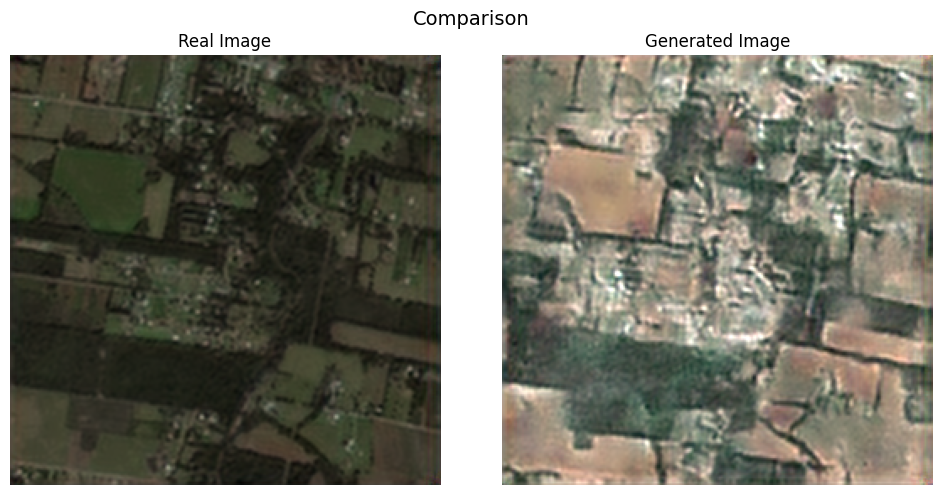}} \\[0.5cm]
    \subfigure[]{\includegraphics[width=0.3\textwidth]{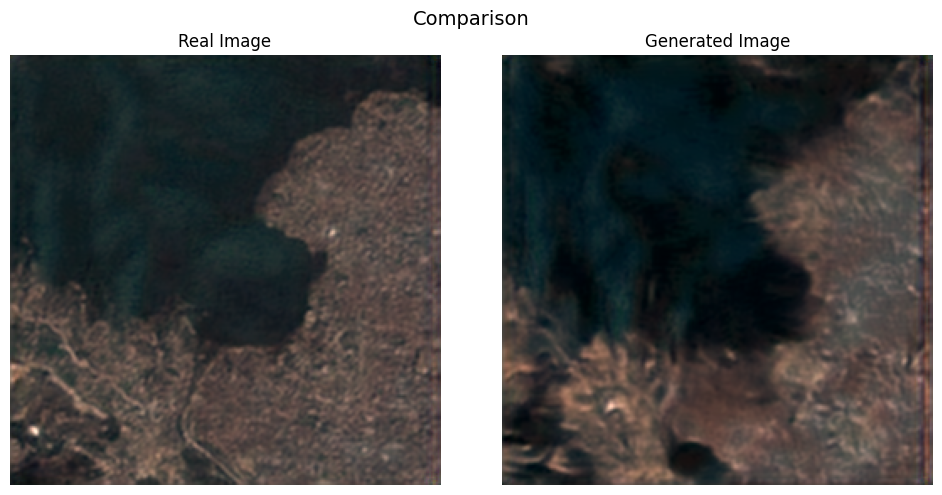}}
    \subfigure[]{\includegraphics[width=0.3\textwidth]{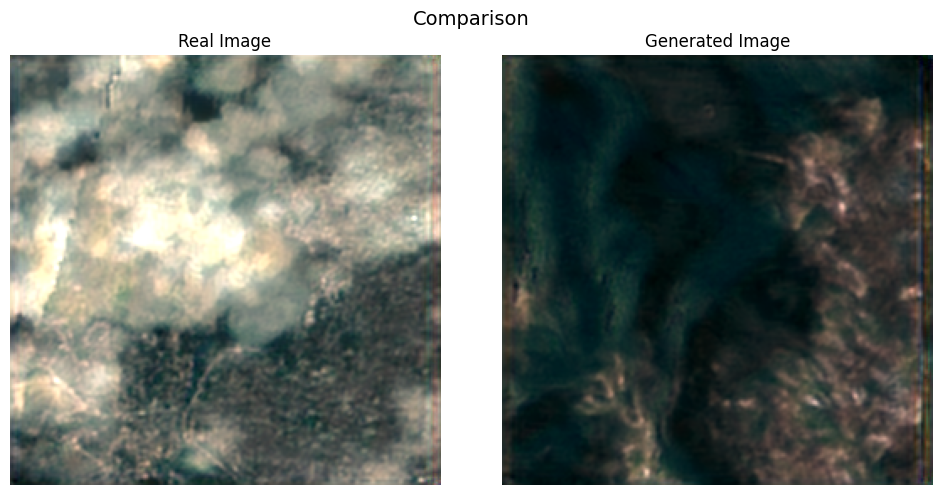}}
    \subfigure[]{\includegraphics[width=0.3\textwidth]{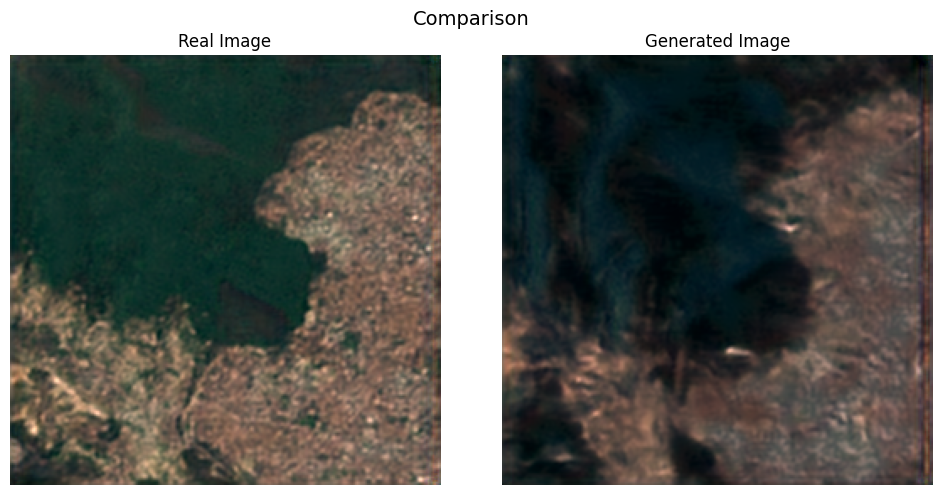}}

    \subfigure[]{\includegraphics[width=0.3\textwidth]{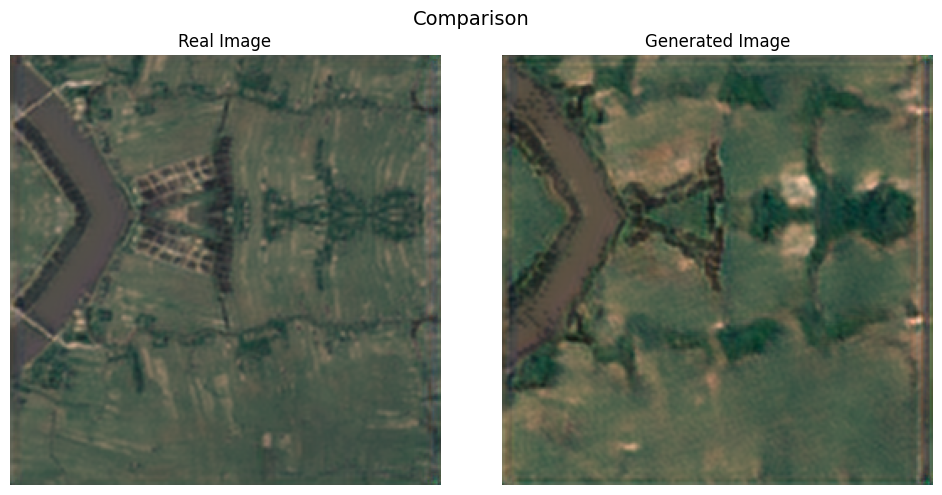}}
    \subfigure[]{\includegraphics[width=0.3\textwidth]{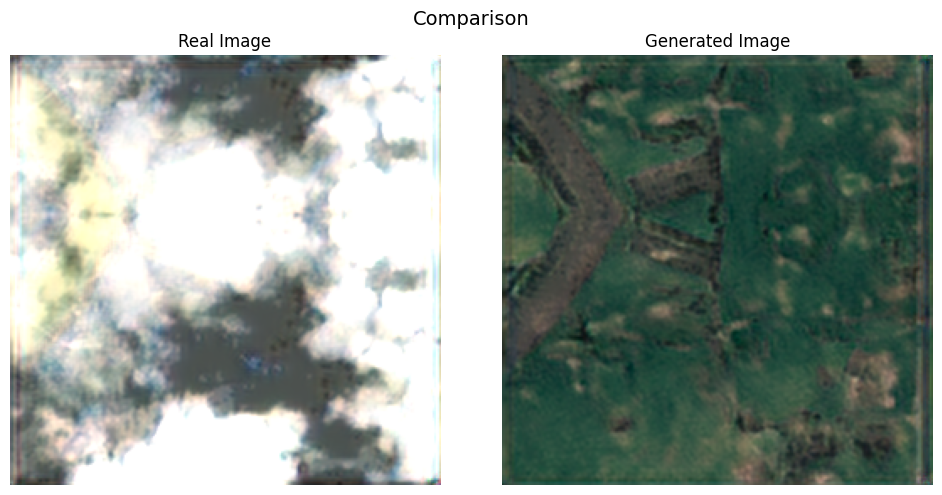}}
    \subfigure[]{\includegraphics[width=0.3\textwidth]{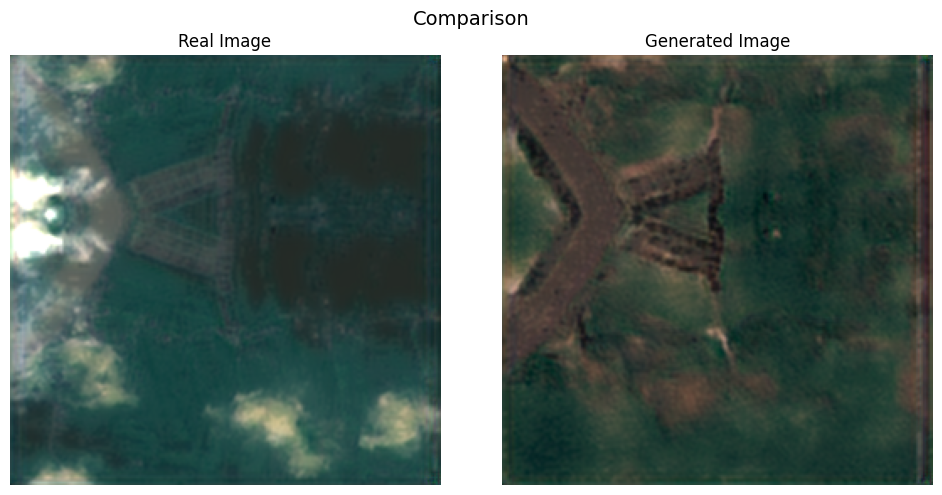}}

    \subfigure[]{\includegraphics[width=0.3\textwidth]{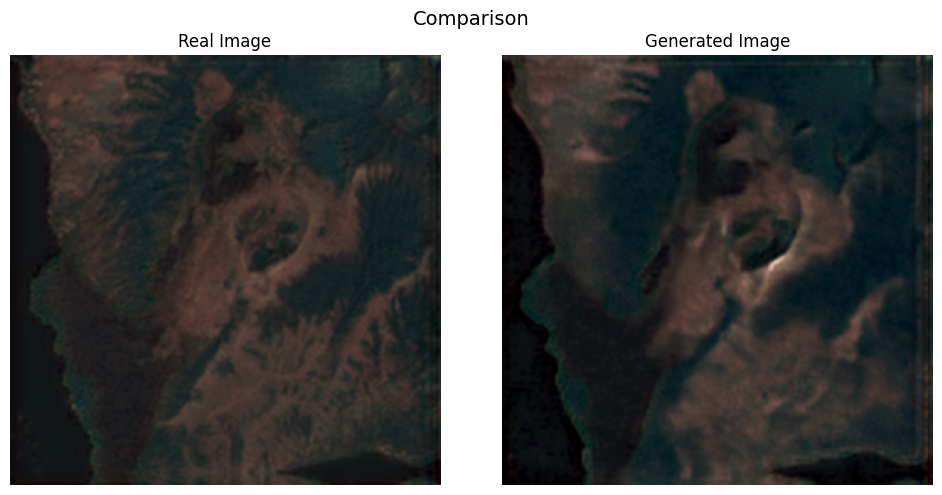}}
    \subfigure[]{\includegraphics[width=0.3\textwidth]{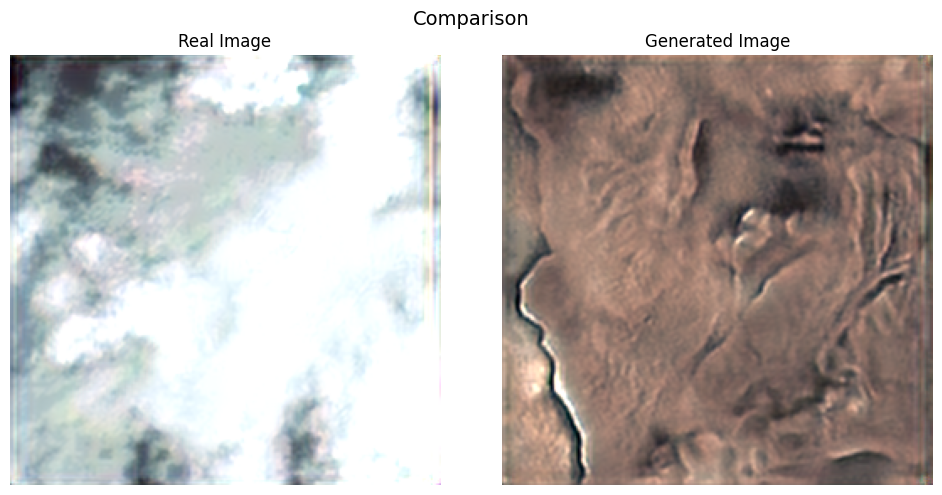}}
    \subfigure[]{\includegraphics[width=0.3\textwidth]{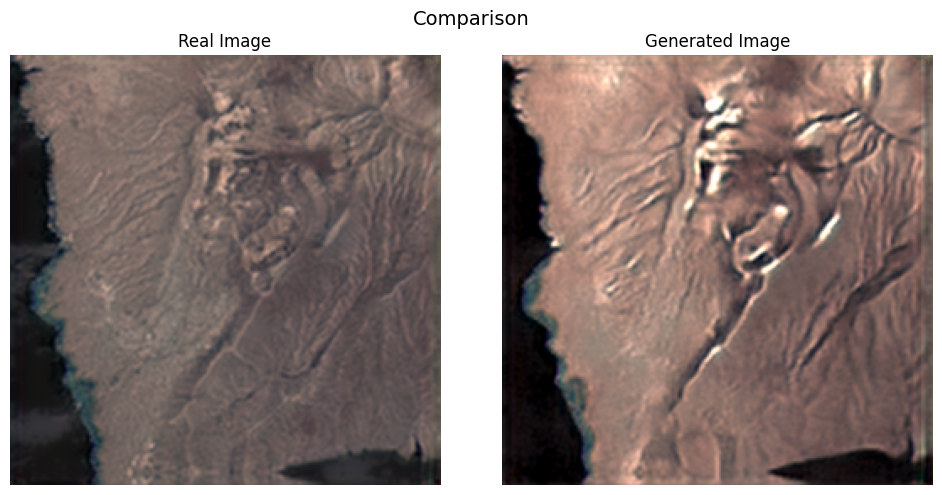}}
    
    \caption{
Comparison of optical (RGB) images for five real-life disaster events. Each row corresponds to a different disaster: Amazon fire, Hurricane Harvey, Nepal flood, Cyclone Remal, and the Taal Volcano eruption. Each row shows three temporal stages: before (first column), during (second column), and after (third column) the event. Within each cell, the left image is the original satellite observation (often cloud-affected during disasters), and the right image is the corresponding reconstruction generated by our model. \textbf{a, d, g, j, m}: before images; \textbf{b, e, h, k, n}: during images; \textbf{c, f, i, l, o}: after images.
}

    \label{fig:disaster_RGB_comparison}
\end{figure}

Fourthly, we examine the impact of Cyclone Remal \cite{cyclone_remal} in 2024. Severe Cyclonic Storm Remal struck West Bengal and Bangladesh in May 2024, causing sustained winds of up to 135 km/h and resulting in at least 85 fatalities. The cyclone disrupted power for around 30 million people in Bangladesh and caused widespread damage in the affected coastal regions. The results from our model for this event are presented in the fourth row of Fig.~\ref{fig:disaster_RGB_comparison} (main text) and Fig.~\ref{fig:disaster_NDVI_comparison} and Fig.~\ref{fig:disaster_NDWI_comparison} (supplementary materials). Fig.~\ref{fig:disaster_RGB_comparison}.j (main text) and Fig.~\ref{fig:disaster_NDVI_comparison}.j and Fig.~\ref{fig:disaster_NDWI_comparison}.j (supplementary materials) present the optical (RGB), NDVI, and NDWI images of the area before the cyclone. These images clearly delineate the pre-disaster landscape, with NDVI and NDWI effectively capturing healthy vegetation and normal water content.  Fig.~\ref{fig:disaster_RGB_comparison}.k (main text) and Fig.~\ref{fig:disaster_NDVI_comparison}.k and Fig.~\ref{fig:disaster_NDWI_comparison}.k (supplementary materials) present the corresponding images during the cyclone, highlighting the inundation and disruption to vegetation. NDVI and NDWI values have declined noticeably, reflecting vegetation loss and increased surface water coverage. A significant change in land cover and persistent water-logged areas can be observed in the generated images on the right, which are not clearly visible in the original images on the left due to cloud cover. Finally, Fig.~\ref{fig:disaster_RGB_comparison}.l (main text) and Fig.~\ref{fig:disaster_NDVI_comparison}.l and Fig.~\ref{fig:disaster_NDWI_comparison}.l (supplementary materials) show the area after the cyclone. 

\subsubsection{Volcanic Eruption}
Finally, we examine the impact of the Taal volcanic eruption \cite{taal_eruption_2020_2022} in the Philippines in 2020. The Taal Volcano eruption in January 2020 disrupted daily life for millions, causing ashfall across Metro Manila and nearby provinces that worsened air quality and led to widespread evacuations. The event forced residents to leave their homes and affected transportation, health, and local economies in the region. The results obtained from our model for this event are as follows: the first row in .The Fig.~\ref{fig:disaster_RGB_comparison}.m (main text) and Fig.~\ref{fig:disaster_NDVI_comparison}.m and Fig.~\ref{fig:disaster_NDWI_comparison}.m (supplementary materials) presents the optical (RGB), NDVI, and NDWI images of the area before the eruption. These images clearly depict the pre-eruption landscape, with NDVI and NDWI capturing healthy vegetation and water distribution. Fig.~\ref{fig:disaster_RGB_comparison}.n (main text) and Fig.~\ref{fig:disaster_NDVI_comparison}.n and Fig.~\ref{fig:disaster_NDWI_comparison}.n (supplementary materials) present the corresponding images during the eruption, highlighting the ash cover, vegetation damage, and alterations in moisture levels. The NDVI and NDWI values decline substantially, reflecting vegetation stress and disruption of surface water patterns. Significant landscape changes can be observed in the generated images on the right, which are not clearly visible in the original images on the left due to heavy cloud and ash cover. Finally, Fig.~\ref{fig:disaster_RGB_comparison}.o (main text) and Fig.~\ref{fig:disaster_NDVI_comparison}.o and Fig.~\ref{fig:disaster_NDWI_comparison}.o (supplementary materials) show the area after the eruption.

Thus, our model was able to generate crucial optical, NDWI, and NDVI representations under different conditions across various disasters. These visualizations demonstrate the utility of our model in monitoring and analyzing the progression and aftermath of various disasters.

\subsection{Future Applications}

Although we have trained CloudBreaker for environments on Earth, its applications need not be limited to our planet. By extending its scope, we can adapt it for use on other celestial bodies, opening the door to uncovering some of the greatest mysteries in space exploration. For instance, it could potentially help us visualize what lies beneath the thick, acidic clouds of Venus  by first training on data from various other planets and then fine-tuning on those with similar features. The potential for CloudBreaker in extraterrestrial exploration is immense. It can be applied to other situations where cloud cover is persistent all around like cloud forest. But for this training in similar environment before applying it will be neccessary. We would also suggest to use inference steps of 1000 or higher for better translation of images as this would be uncharted territory.

\section{Conclusion}

We have successfully developed a method capable of removing clouds from Sentinel-2 images by leveraging Sentinel-1 data to reconstruct multi-spectral imagery. These reconstructed multi-spectral images enable the retrieval of optical information as well as the computation of valuable vegetation indices, such as NDWI and NDVI, which are useful for a wide range of practical applications. This work represents a significant step toward overcoming the limitations imposed by cloud cover in remote sensing. Moreover, our novel multi-stage training process facilitates easier learning of the translation path, which can be generalized to other flow matching frameworks. Additionally, our introduction of cosine scheduling into flow matching offers a new tool
for use in generative models. Future research can explore alternative non-linear scheduling schemes to further enhance reconstruction quality.


\begin{table*}[ht]
\centering
\begin{tabularx}{\textwidth}{l c c c c c c c c c}
\toprule
Method & Steps & \shortstack{RGB\\FID \\ $\downarrow$} & \shortstack{MSE$_\text{latent}$ \\ $\downarrow$} & \shortstack{R$^2_\text{latent}$ \\ $\uparrow$} & \shortstack{RGB \\ SSIM \\ $\uparrow$} & \shortstack{RGB \\ LPIPS \\ $\downarrow$} & \shortstack{NDVI \\ SSIM \\ $\uparrow$} & \shortstack{NDWI \\ SSIM \\ $\uparrow$} \\
\midrule
FM-Cosine & 100  & 0.7432 & 0.003814 & 0.4969 & 0.6346 & 0.2719 & 0.6156 & 0.6874 \\
FM-Cosine & 1000 & 0.6481 & 0.004292 & 0.4321 & 0.6193 & 0.2791 & 0.5979 & 0.6722 \\
FM-Expo $k=1.0$ & 100  & 3.0730 & 0.002453 & 0.6721 & 0.5711 & 0.3361 & 0.5459 & 0.5925 \\
FM-Expo $k=1.0$ & 1000 & 2.2845 & 0.002566 & 0.6560 & 0.5619 & 0.3347 & 0.5293 & 0.5812 \\
FM-Expo $k=1.5$ & 100  & 3.2090 & 0.002488 & 0.6671 & 0.5707 & 0.3321 & 0.5529 & 0.5935 \\
FM-Expo $k=1.5$ & 1000 & 2.7888 & 0.002545 & 0.6588 & 0.5670 & 0.3283 & 0.5467 & 0.5886 \\
FM-Expo $k=2.0$ & 100  & 1.4388 & 0.002620 & 0.6473 & 0.5403 & 0.3447 & 0.4823 & 0.5384 \\
FM-Expo $k=2.0$ & 1000 & 0.6840 & 0.002944 & 0.6012 & 0.5104 & 0.3572 & 0.4267 & 0.4912 \\
FM-Expo $k=2.5$ & 100  & 2.3030 & 0.002613 & 0.6486 & 0.5578 & 0.3321 & 0.5412 & 0.5803 \\
FM-Expo $k=2.5$ & 1000 & 2.0311 & 0.002667 & 0.6408 & 0.5541 & 0.3309 & 0.5376 & 0.5760 \\
FM-Linear & 100  & 2.2573 & 0.002892 & 0.6077 & 0.5148 & 0.3756 & 0.5062 & 0.5257 \\
FM-Linear & 1000 & 1.6853 & 0.003182 & 0.5661 & 0.4985 & 0.3735 & 0.4900 & 0.5072 \\
BBDM & -- & 5.9641 & 0.003352 & 0.5435 & 0.4388 & 0.4286 & 0.4504 & 0.5013 \\
Pix2Pix & -- & 3.9371 & 0.005543 & 0.2094 & 0.3160 & 0.4205 & 0.2995 & 0.2948 \\
CycleGAN & -- & 6.9654 & 0.003898 & 0.4609 & 0.4733 & 0.4278 & 0.3774 & 0.4386 \\
Ahmed et al.~\cite{sakib2025light} & -- & -- & -- & -- & -- & -- & -- & 0.4226 \\
\bottomrule
\end{tabularx}
\caption{Performance metrics for different schedulers and number of steps. 
"FM-" denotes Flow Matching with different interpolation strategies and step counts. 
"BBDM" refers to the Brownian Bridge Diffusion Model. 
"Pix2Pix" and "CycleGAN" are supervised and cycle-consistent baselines, respectively, without explicit step-based sampling. Additionally, we evaluated Ahmed et al.'s model \cite{sakib2025light} on the NDWI, which is its only output.
}
\label{tab:scheduler_metrics}
\end{table*}

\begin{figure}[htbp]
    \centering
    \subfigure[]{\includegraphics[width=0.3\textwidth]{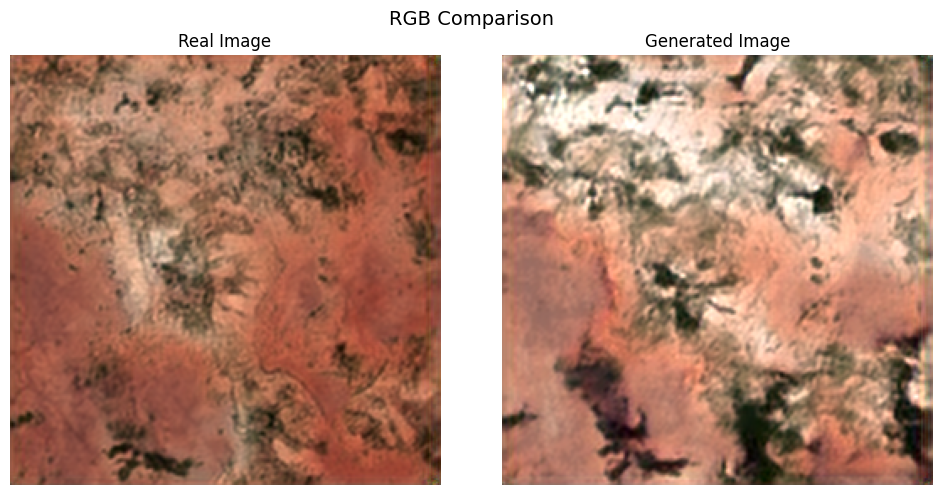}}
    \subfigure[]{\includegraphics[width=0.3\textwidth]{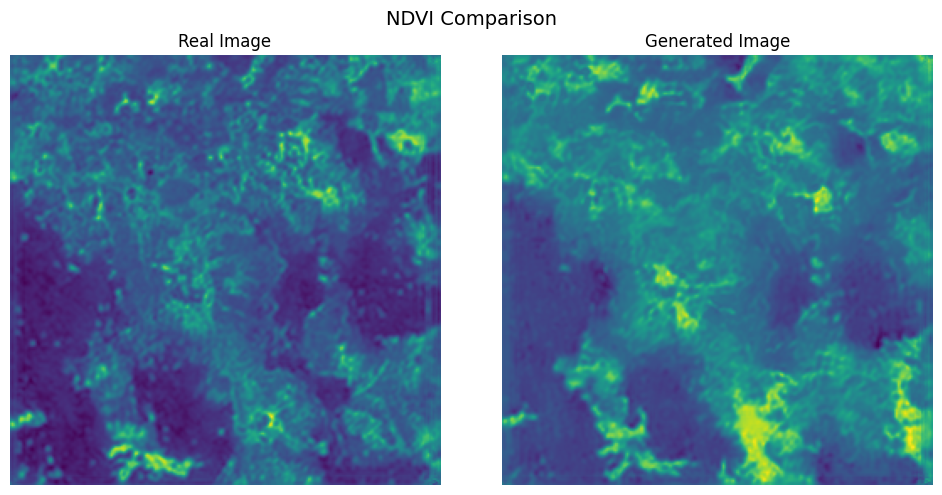}}
    \subfigure[]{\includegraphics[width=0.3\textwidth]{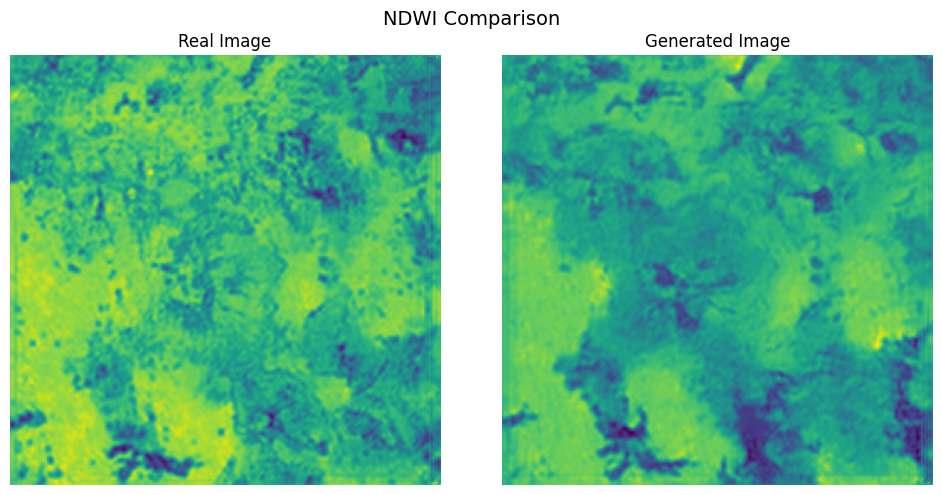}} \\
    \subfigure[]{\includegraphics[width=0.3\textwidth]{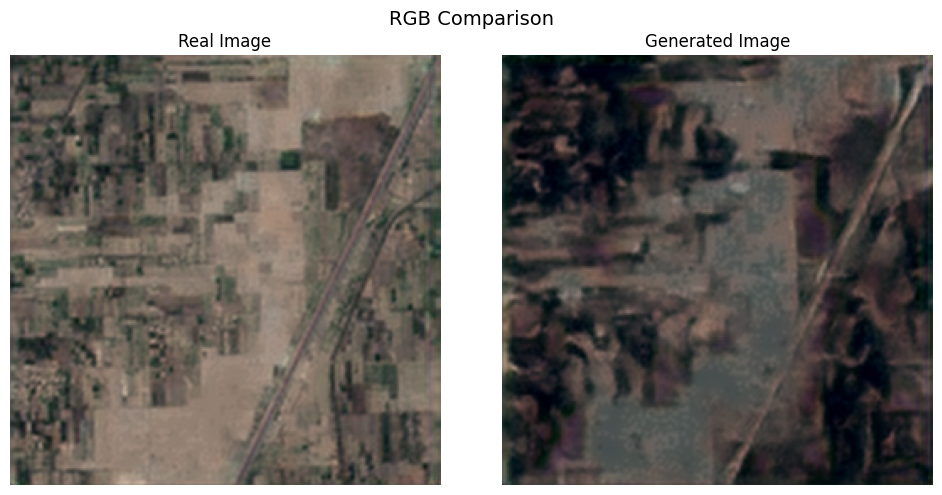}}
    \subfigure[]{\includegraphics[width=0.3\textwidth]{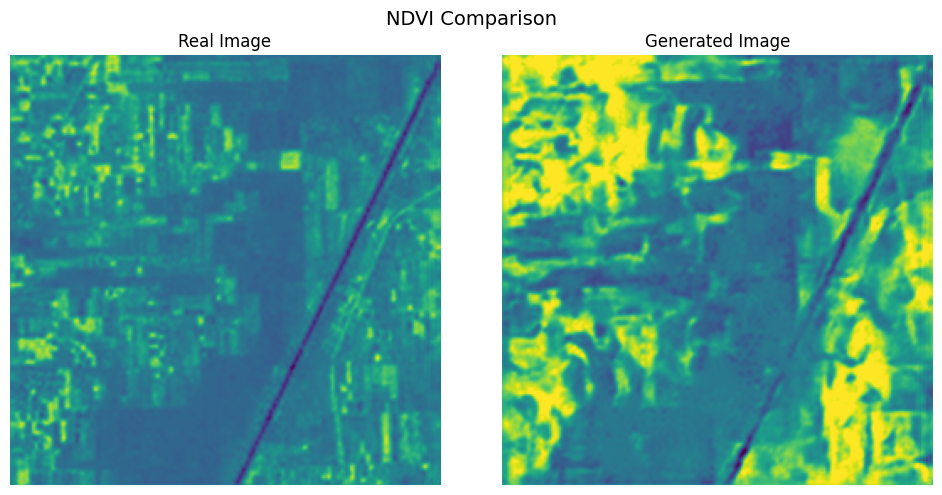}}
    \subfigure[]{\includegraphics[width=0.3\textwidth]{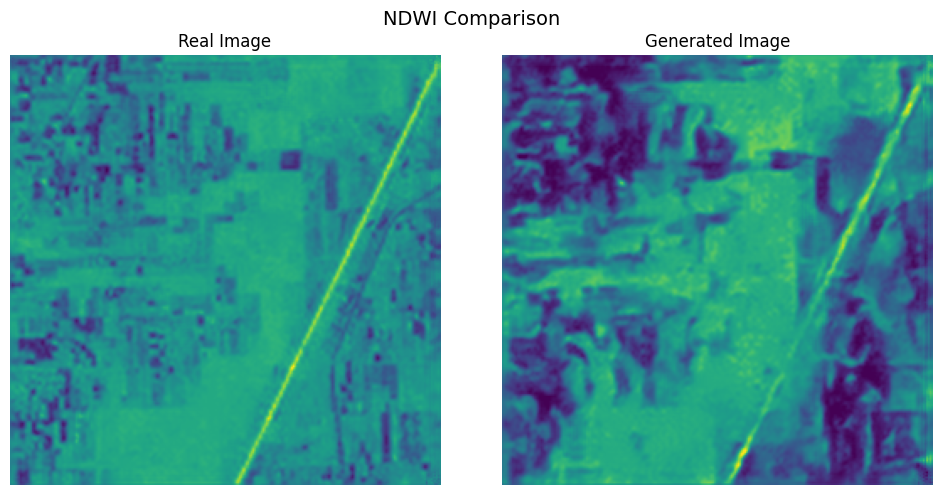}} \\
    \subfigure[]{\includegraphics[width=0.3\textwidth]{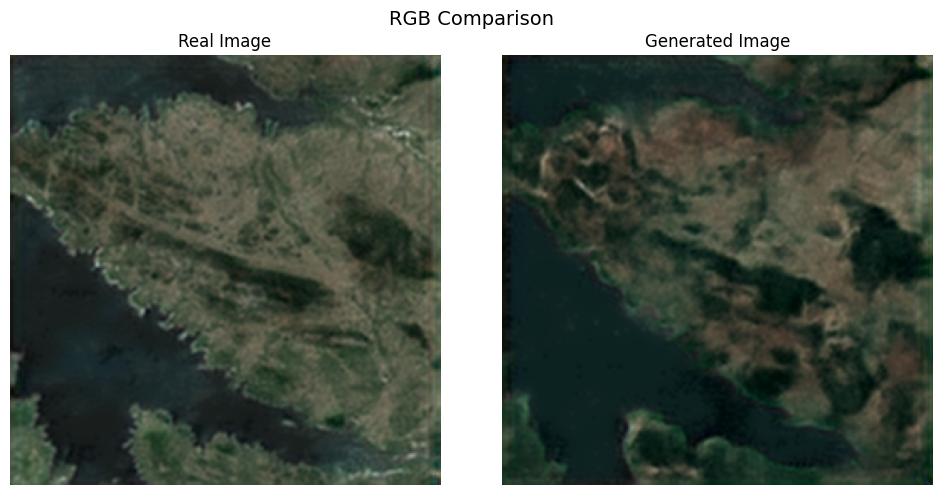}}
    \subfigure[]{\includegraphics[width=0.3\textwidth]{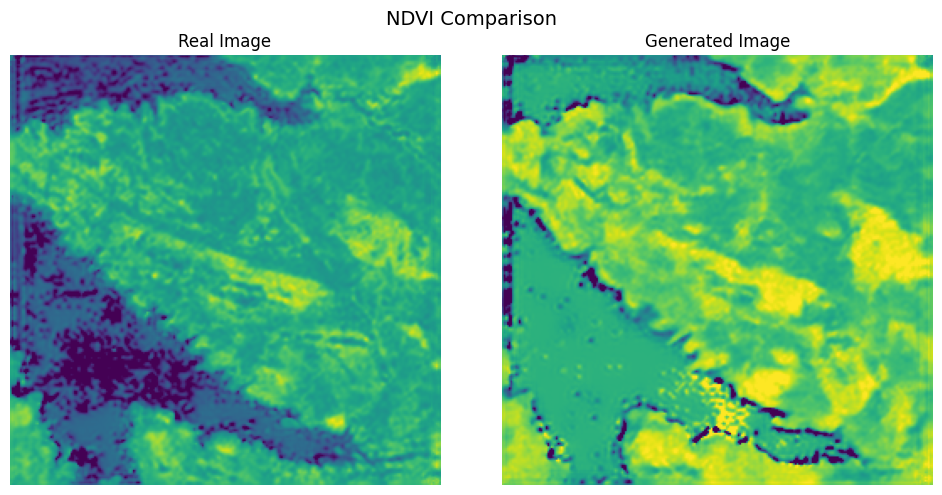}}
    \subfigure[]{\includegraphics[width=0.3\textwidth]{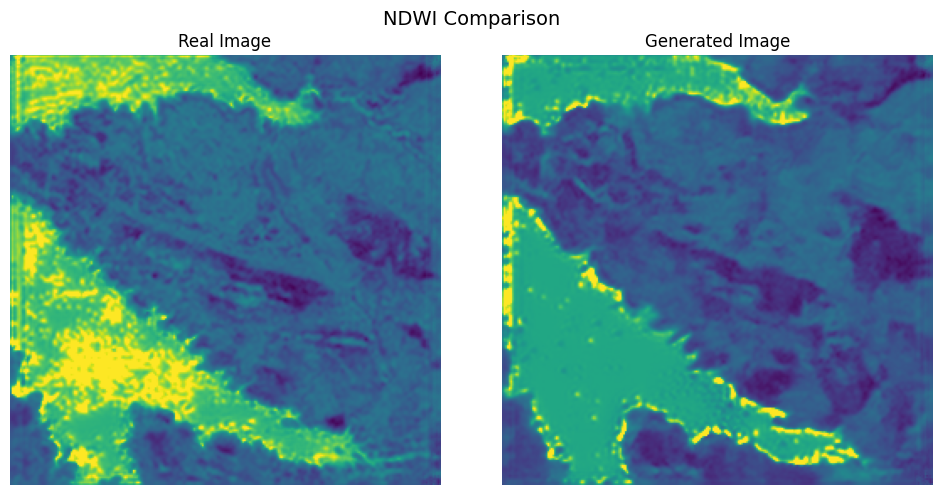}} \\
    \subfigure[]{\includegraphics[width=0.3\textwidth]{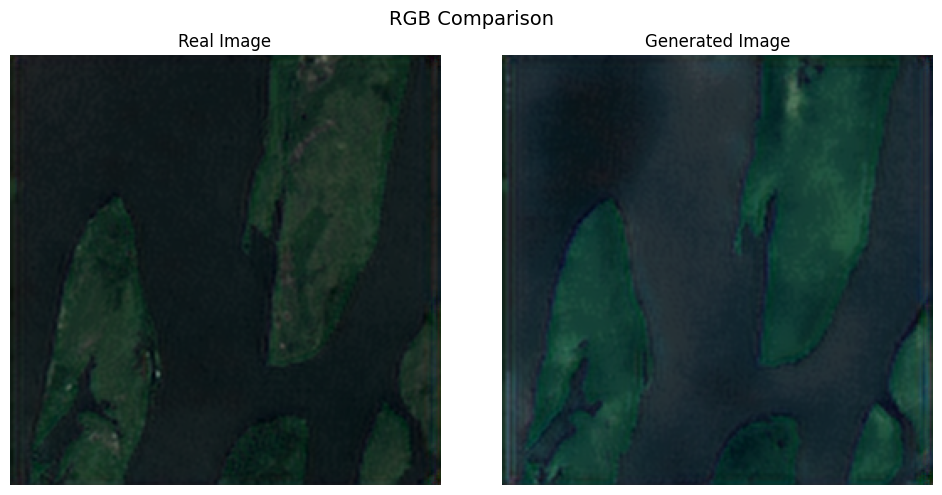}}
    \subfigure[]{\includegraphics[width=0.3\textwidth]{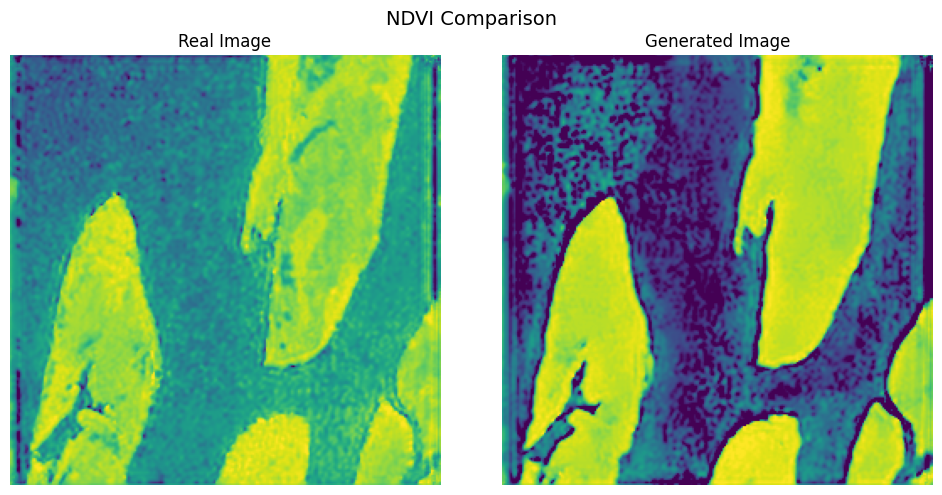}}
    \subfigure[]{\includegraphics[width=0.3\textwidth]{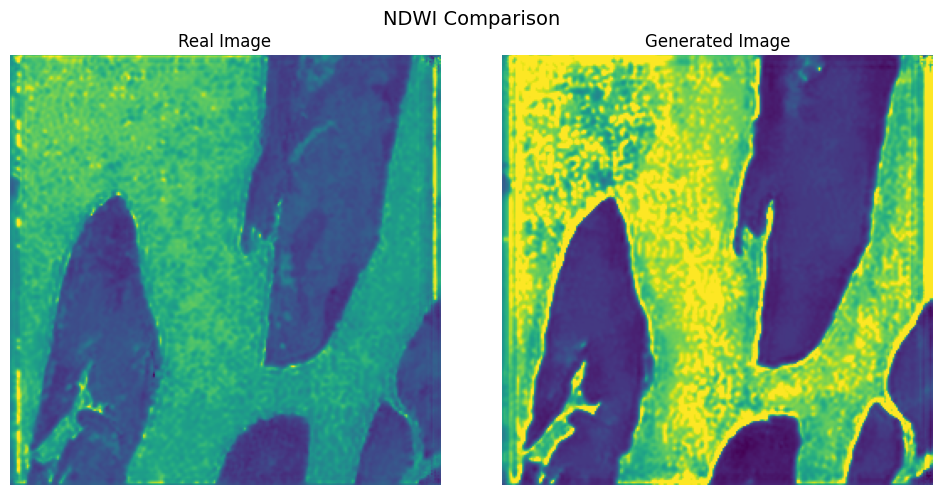}}
    
    \caption{\textbf{a, d, g, j:} Comparison between real (left) and generated (right) RGB images. \textbf{b, e, h, k:} Comparison between real (left) and generated (right) NDVI images. \textbf{c, f, i, l:} Comparison between real (left) and generated (right) NDWI images.}

    \label{fig:cloud_free_images}
\end{figure}

\FloatBarrier

\section{Author Contribution}

\textbf{Saleh Sakib Ahmed:} Significantly contributed to idea formulation, method development, model architecture design, and paper writing.\\
\textbf{Sara Nowreen:} Provided guidance and support, paper writing, and review.\\
\textbf{M. Sohel Rahman:} Provided guidance and support, paper writing, and review.

\section{Acknowledgement}
The authors thank Md. Sabbir Hossain from IWFM, BUET, for providing the Sentinel-1 and Sentinel-2 images of various disasters used in our analysis.

\bibliography{sn-bibliography}

\clearpage
\appendix
\section*{Supplementary Material}
\renewcommand{\thesection}{S\arabic{section}}
\renewcommand{\thesubsection}{S\arabic{section}.\arabic{subsection}}

\section{Methods and Materials}\label{sec:methods}

\subsection{Datasets}

In this research, we utilized the dataset curated by Ahmed et al.~\cite{sakib2025light}, which was derived from the Cloud to Street - Microsoft Flood and Clouds Dataset, made publicly available by the Radiant Earth Foundation~\cite{cloud2024}. The dataset consists of 900 paired image chips from Sentinel-1 and Sentinel-2 satellites, collected from 18 major flood events across different regions of the world.

Each Sentinel-1 chip includes two radar backscatter bands: VV (vertical transmit and receive) and VH (vertical transmit and horizontal receive). The values are provided in decibel (dB) units, and typically range between \textbf{-25 dB and 0 dB}, although extreme values may be present.

Sentinel-2 chips consist of 13 spectral bands, including the Red, Green, Blue (RGB), and Near-Infrared (NIR) bands. The pixel values represent surface reflectance and are originally scaled as unsigned integers ranging from 0 to a maximum of 10,000. However, in practice, most reflectance values fall below \textbf{5,000}, particularly in non-saturated, non-cloudy regions.

All image chips are of spatial resolution 512 $\times$ 512 pixels and include scenes captured under both cloud-free and cloudy conditions. Additionally, the dataset includes high-quality binary masks for surface water and cloud coverage. We used the cloud cover mask to find cloud-free images.

\subsection{Code and Data Availability}

The dataset used in this study is available at:  
\url{https://www.kaggle.com/datasets/sakibahmed91/cloud2street-dataset}

The data for the various case studies of disasters are available at: \url{https://drive.google.com/drive/folders/1Xt9tpl72Idw6__lnqvTJRiUSM4iII8KJ}

The weight of the best-performing CloudBreaker model (with cosine learning rate scheduling) can be accessed at:  
\url{https://www.kaggle.com/datasets/sakibahmed91/weights-of-cloudbreaker-larger-version}

The weights of the Scaler and VQ models for both Sentinel-1 and Sentinel-2 are available at:  
\url{https://www.kaggle.com/datasets/sakibahmed91/vq-model-and-scaler-model-weights}

All source code, along with the links to the model weights, is available in the GitHub repository:  
\url{https://github.com/bojack-horseman91/Cloudbreaker-Large/tree/main}

\subsection{Metric Used} \label{subsec:metric_used}

We evaluated the realism of generated RGB images using Fréchet Inception Distance (FID) \cite{dowson1982frechet,heusel2017gans}, Structural Similarity Index (SSIM) \cite{nilsson2020understanding}, and Learned Perceptual Image Patch Similarity (LPIPS) \cite{zhang2018unreasonable}. FID (range: [0, $\infty$]; lower is better) measures the distance between feature distributions of real and generated images using a pretrained Inception network, effectively capturing both image quality and diversity. SSIM (range: [0, 1]; higher is better) assesses structural similarity based on luminance, contrast, and structure, but is sensitive to pixel-level differences. LPIPS (range: [0, 1]; lower is better) uses deep features from pretrained networks to estimate perceptual similarity and aligns well with human judgment. Among these, FID is the most preferred for generative image evaluation due to its strong correlation with human perception and its ability to capture both fidelity and diversity \cite{arabboev2024comprehensive}. As previously mentioned, the FID score ranges from 0 to infinity, where lower values indicate better quality and higher similarity to real images. A score closer to 0 means the generated images are more realistic and diverse. We have used Mean Squared Error (MSE) and $R^2$ score to evaluate the similarity of the translated latent space of Sentinel-2. MSE gives us the mean of pixel wise error whereas $R^2$ score gives the alignment between target and output latent space. Although $R^2$ score gives alignment, it should be taken with a grain of salt as it fails to capture the nonlinear, perceptual, and structural aspects that are crucial for evaluating generated images \cite{quantics2022r2}. For evaluating the single channel NDWI and NDVI we used SSIM.

\subsection{Data Preprocessing}

The original \(512 \times 512\) pixel chips were divided into 16 smaller \(256 \times 256\) chips, increasing the number of Sentinel-1 and Sentinel-2 pairs from 900 to 3600. After applying cloud masks to exclude cloud-covered samples, 1679 cloud-free chips remained for training and evaluation. Each chip was normalized to the \([0, 1]\) range. From this dataset, 70\% was used for training, 20\% for testing, and 10\% for validation.

\textbf{Sentinel-1} data includes two polarization bands: VV (vertical transmit/receive) and VH (vertical transmit, horizontal receive), each represented as a \(256 \times 256\) 2D array. These are stacked along the channel dimension to form a 3D input of shape \((256, 256, 2)\). \textbf{Sentinel-2} data includes four spectral bands: Red, Green, Blue (RGB), and Near Infrared (NIR), also in the form of \(256 \times 256\) arrays. These are combined to form a 3D array of shape \((256, 256, 4)\).

\subsection{Image Scaling Procedure}

To ensure numerical stability and improve model training performance, all satellite images were normalized using a custom scaling approach implemented via the \texttt{ImageScaler} class. This scaler performs per-channel normalization based on either fixed min/max values or percentile ranges derived from the data itself.

The scaling process operates as follows:
\begin{itemize}
    \item \textbf{Channel-wise Scaling:} The image tensor is reshaped such that each channel is treated independently. This allows for normalization specific to the distribution of each band (channel), which is particularly important when different bands have different physical ranges or value distributions.
    \item \textbf{Percentile-based Normalization:} Instead of using global minimum and maximum values—which can be sensitive to outliers—the scaler computes the \texttt{pmin}\% and \texttt{pmax}\% percentiles (e.g., 1st and 98th) from all pixel values in each channel. These values are then used as the lower and upper bounds for scaling.
    \item \textbf{Linear Transformation:} Each channel is linearly scaled using the formula:
    \[
    \text{scaled} = \frac{\text{value} - \text{pmin}}{\text{pmax} - \text{pmin} + \varepsilon}
    \]
    where $\varepsilon$ is a small constant (e.g., $1 \times 10^{-6}$) added to avoid division by zero.
    \item \textbf{No Clipping:} The implementation intentionally avoids clipping the scaled values to $[0, 1]$ to preserve useful gradients and prevent artificial saturation during training. But we do clip the data during the practical application so that unexpected high data do not cause big issue. We first observe the data distribution and if we find data are getting very high which happens in Sentinel-2 when clouds are present in the images we clip the data to 5000.
\end{itemize}

We applied this procedure separately to Sentinel-1 and Sentinel-2 datasets. For Sentinel-1 images, we used the 0.1\textsuperscript{th} and 99.9\textsuperscript{th} percentiles as the scaling range to account for its typically higher dynamic range. For Sentinel-2 images, we used the 1\textsuperscript{st} and 98\textsuperscript{th} percentiles, which proved effective in suppressing cloud and shadow-related outliers.

This percentile-based normalization retains the meaningful structure in each channel while minimizing the influence of extreme values. It also standardizes the input for deep learning models, which often assume feature values lie in a similar range across channels.

\textbf{Data Augmentation:} To increase the diversity of the training data, each original image pair was duplicated with augmentations applied to the copy, effectively doubling the training dataset size. The applied augmentations included random horizontal flip, vertical flip, and rotation up to 20 degrees. To ensure alignment between Sentinel-1 and Sentinel-2 images, both were concatenated along the channel dimension before augmentation and then split back into their respective components afterward.

\subsection{Latent Space}

As mentioned previously, our objective is to transform the input distribution into the target distribution by mapping both into a shared latent space. To achieve this, we employed two separate Vector-Quantized Variational Autoencoders (VQ-VAE) \cite{van2017neural}, one for encoding Sentinel-1 images and another for Sentinel-2 images. Each VQ-VAE encodes the input into a latent representation with 16 channels and spatial dimensions of \(128 \times 128\), resulting in a latent space of shape \(128 \times 128 \times 16\). The models were implemented using the Hugging Face library and configured with one downsampling and one upsampling layer, each with 32 channels. The downsampling reduces the spatial resolution by a factor of 2, projecting the input into the 16-channel latent space, while the upsampling reconstructs the original dimensions.

This configuration ensures that both the 2-channel Sentinel-1 and 4-channel Sentinel-2 inputs are encoded into the same latent dimensionality, enabling computation of a difference vector \(\Delta Z_s = Z_{S_2} - Z_{S_1}\), where \(Z_{S_1}\) and \(Z_{S_2}\) denote the latent representations of Sentinel-1 and Sentinel-2, respectively. This difference vector guides the model in learning the transformation from the source to the target distribution.

\subsection{Architecture of models}

\subsubsection{Model Architecture for VQVAE}

We used two VQ-VAE implemented via the \texttt{VQModel} \cite{huggingface_vqmodel} class from Hugging Face's \texttt{diffusers} library, to encode Sentinel-1 and Sentinel-2 images into a compact latent space of shape $128 \times 128 \times 16$. For Sentinel-1, the model accepted 2 input channels and produced 2 output channels, while for Sentinel-2, it used 4 input and output channels. Both models shared a similar architecture comprising one downsampling block (\texttt{DownEncoderBlock2D}) and one upsampling block (\texttt{UpDecoderBlock2D}), with 7 convolutional blocks configured with output channels set to $(32, 32, 32, 32, 32, 32, 32)$ and 3 layers per block. The number of embeddings in the codebook was set to 1024, and no attention mechanism was used in the mid-block (\texttt{mid\_block\_add\_attention=False}). The models were trained using the Adam optimizer with a learning rate of $1 \times 10^{-4}$ and weight decay values of $1 \times 10^{-10}$ for Sentinel-1 and $1 \times 10^{-13}$ for Sentinel-2. A \texttt{ReduceLROnPlateau} scheduler was applied to reduce the learning rate when the validation loss plateaued, ensuring efficient convergence during training.

\subsubsection{Model Architecture for Translation Model}

The model used for the image translation task is based on the \texttt{UNet2DModel} from the \texttt{diffusers} library of HuggingFace. It is configured with an input channel size of $2 \times \texttt{NUM\_INPUT\_CHANNEL}$ (latent space Sentinel-1 band for conditioning and current iteration translation of Sentinel-1 to Sentinel-2) and outputs \texttt{NUM\_OUTPUT\_CHANNEL} (latent space channel of Sentinel-2). The target spatial resolution of the model is set to the resolution of the latent space \(128 \times 128\). The U-Net architecture utilizes progressive channel scaling in its encoder-decoder structure, with the output channels of each block set as $(128, 128, 256, 512, 512)$.

The downsampling path consists of five blocks in the following order: two \texttt{DownBlock2D} layers without attention, followed by three \texttt{AttnDownBlock2D} layers that integrate self-attention to capture complex spatial dependencies. The upsampling path is symmetric, starting with three \texttt{AttnUpBlock2D} layers followed by two \texttt{UpBlock2D} layers for reconstruction. Each block contains two layers, and both the upsampling and downsampling use \texttt{resnet}-style operations. Group normalization is applied with 32 groups, and dropout is introduced at a rate of 0.1. Attention mechanisms are enabled throughout relevant layers.

The training is performed using the Adam optimizer with a learning rate of $1 \times 10^{-4}$ and a weight decay of $1 \times 10^{-8}$. The learning rate is scheduled using a cosine annealing strategy with warm restarts, where the initial restart period ($T_0$) is 13000 steps, the restart multiplier is 2, and the minimum learning rate is set to $1 \times 10^{-6}$. But the model is trained for 700 epochs.

\subsection{Scheduling Scheme and Training Method for the Translation Process}

As mentioned in Section~\ref{sec:intro}, we employ cosine interpolation, as defined in Eq.~\ref{eq:cosine}, to schedule the translation process in the Vector-Quantized (VQ) latent space of Sentinel-1 and Sentinel-2. The core model is a U-Net architecture (\texttt{UNet2DModel}) that operates entirely within the latent space. At each training step, we generated an intermediate latent code \(x_t\) as a weighted blend of \(Z_{S_1}\) and \(Z_{S_2}\), with the weight \(m_t \in [0, 1]\) determined by a cosine schedule (Eq.~\ref{eq:cosine}). This approach allowed the model to learn smooth transitions from input to output with smaller updates initially and larger steps in the final phase, improving generalization across intermediate representations. 

To further enrich training, we adopted a novel multi-stage training approach that utilizes a threefold sampling strategy per batch, as outlined in Algorithm~\ref{algo:cosine}. The first approach involves training on randomly sampled steps along the \emph{continuous} interpolation path, where \(m_t\) is uniformly drawn from \([0, 1]\), generating a random \(x_t\) from the continous path for each training example. The second is the \emph{discrete} mode, in which an integer \(t \in \{0, \dots, N-1\}\) is sampled and \(m_t = t/N\), corresponding to a fixed step along the scheduled inference trajectory (e.g., 1 of 100 steps). Finally, we emphasize a \emph{boundary focus} mode where we always include \(m_t = 0\), corresponding to pure \(Z_{S_1}\), to strengthen the model’s performance on the most challenging transformation steps encountered early in the interpolation path.

\subsection{Separating Channels to get useful signals}

After obtaining the latent representation of Sentinel-2 from Sentinel-1, following the Algorithm`\ref{algo:inference}, we decode it using the VQModel to reconstruct the Sentinel-2 image. We then separate the RGB channels and the Near-Infrared (NIR) channel. Using Eq.~\ref{eq:NDWI} and Eq.~\ref{eq:NDVI}, we compute the NDWI and NDVI indices, respectively.

\subsection{Fine-Tuning and Practical Application}

Although we have used a globally distributed dataset, as mentioned in the ``\textit{Datasets}'' subsection, it does not encompass every possible environment worldwide. Therefore, for practical applications, it is essential to fine-tune the model to the specific environmental conditions of the target location. The fine-tuning process begins with collecting cloud-free images of the area of interest. While collecting these images, special attention must be paid to the range and distribution of pixel values, as outlined in the ``\textit{Datasets}'' subsection. Although it is possible to fine-tune the model using only pre-event images taken at close temporal proximity, better results can be achieved by including cloud-free images from various environmental and seasonal conditions to enhance the model's reconstruction ability. For instance, in the case of the Taal volcanic eruption, we fine-tuned the model using two sets of images from different time points (before and after the eruption) to capture the nature of the volcano under varying conditions. In contrast, for other disasters, such as floods or hurricanes, we fine-tuned the model using only cloud-free pre-event images. After preparing the localized dataset, the model should be fine-tuned on this data. We recommend fine-tuning the U-Net translation model for 100 to 500 epochs to achieve optimal results; however, users should closely monitor validation metrics to determine the ideal stopping point. Once fine-tuned, the model can be deployed in the target region to support applications such as disaster response, change detection, or cloud-cover compensation. If cloud-free pre-event images are not available, users may still generate approximate outputs using extended inference (e.g., 1000 steps or more), though these should be considered rough estimations rather than precise reconstructions. Finally, if the model is to be deployed for production in a specific location, we strongly recommend training it under diverse conditions and scenarios to ensure robust and reliable performance.

\begin{figure}[htbp]
    \centering
    \subfigure[]{\includegraphics[width=0.3\textwidth]{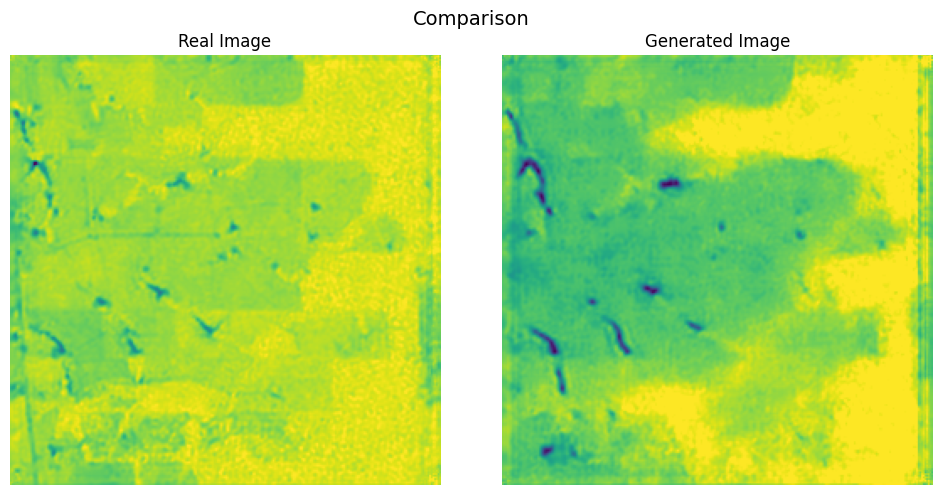}}
    \subfigure[]{\includegraphics[width=0.3\textwidth]{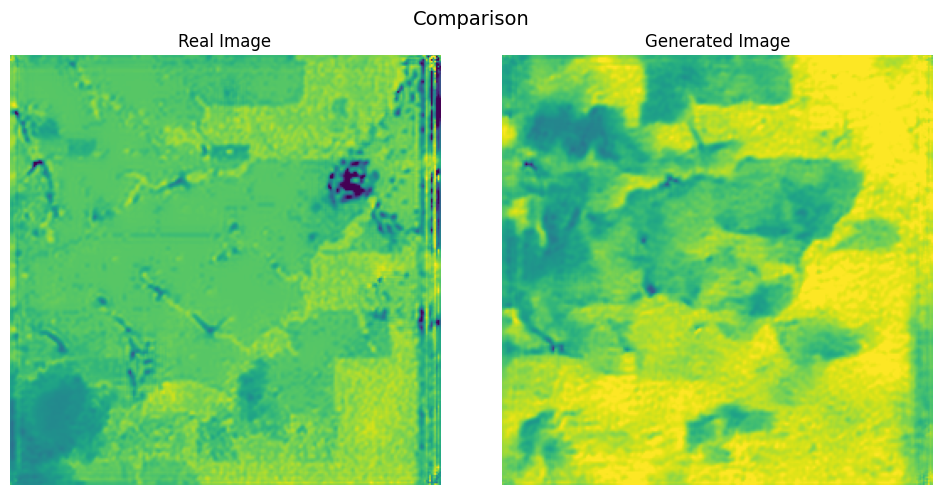}}
    \subfigure[]{\includegraphics[width=0.3\textwidth]{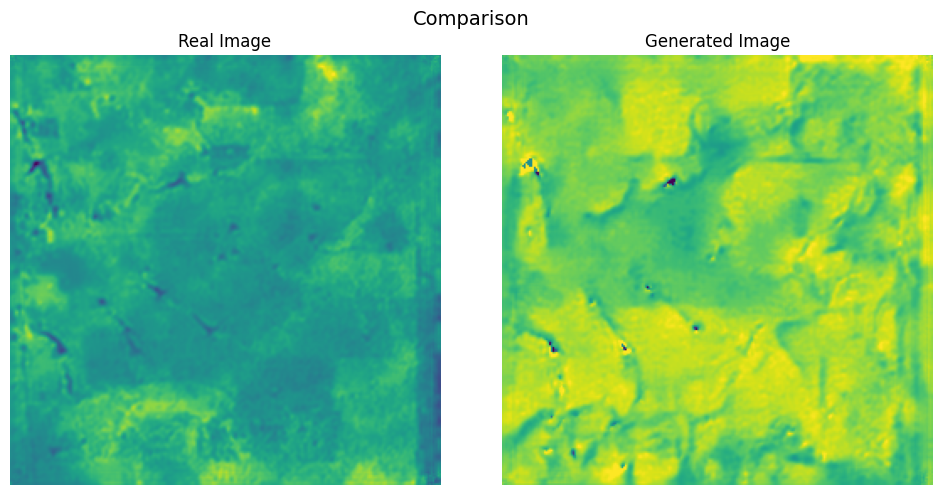}} \\[0.5cm]
    \subfigure[]{\includegraphics[width=0.3\textwidth]{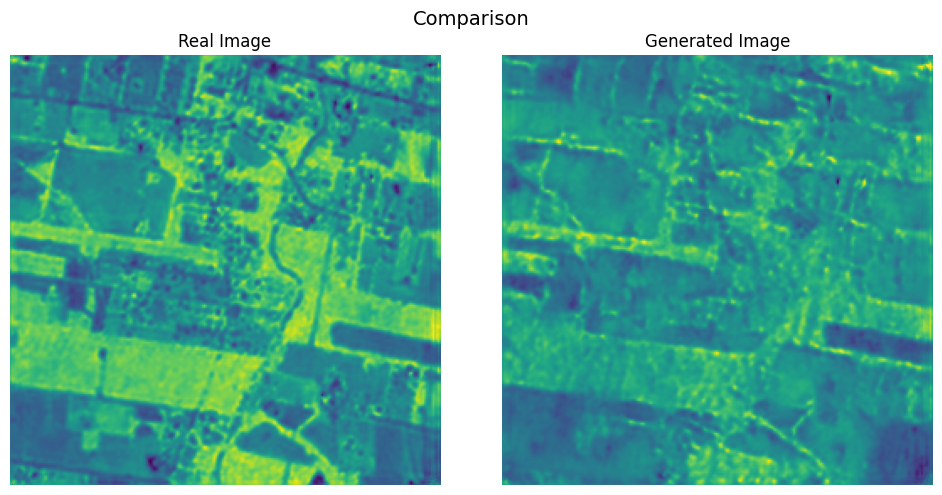}}
    \subfigure[]{\includegraphics[width=0.3\textwidth]{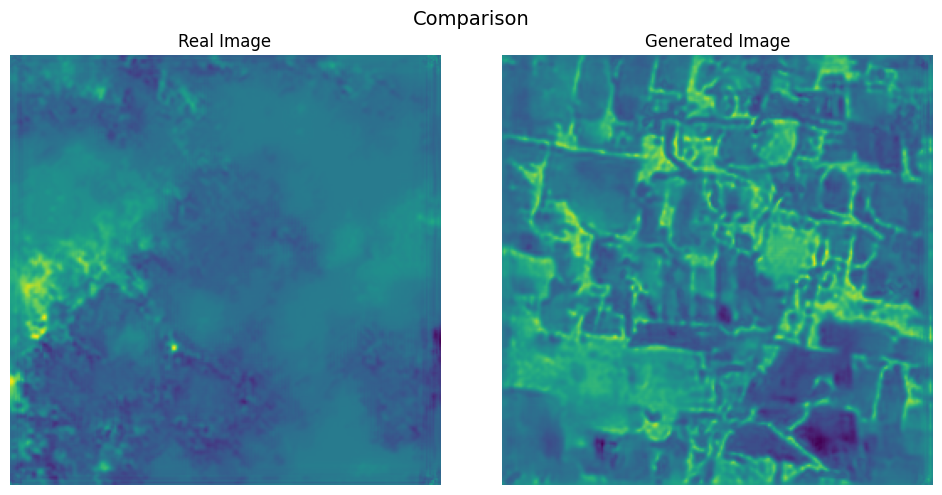}}
    \subfigure[]{\includegraphics[width=0.3\textwidth]{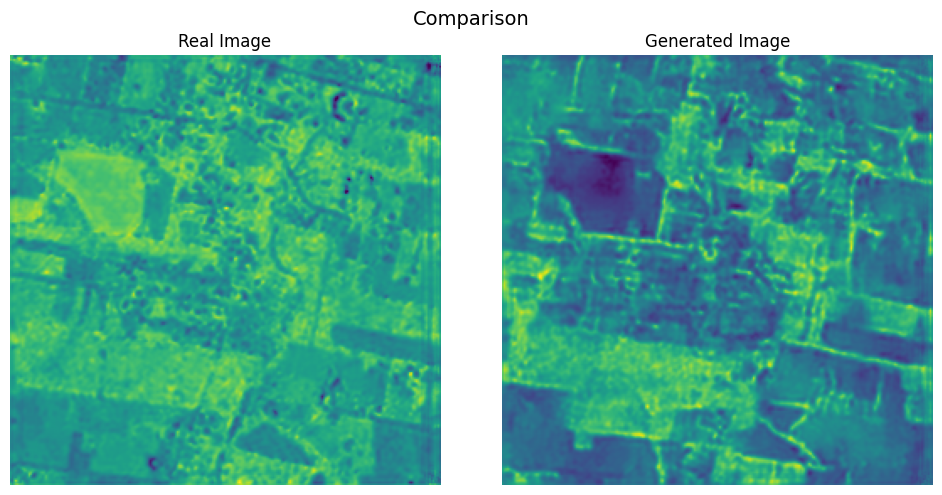}} \\[0.5cm]
    \subfigure[]{\includegraphics[width=0.3\textwidth]{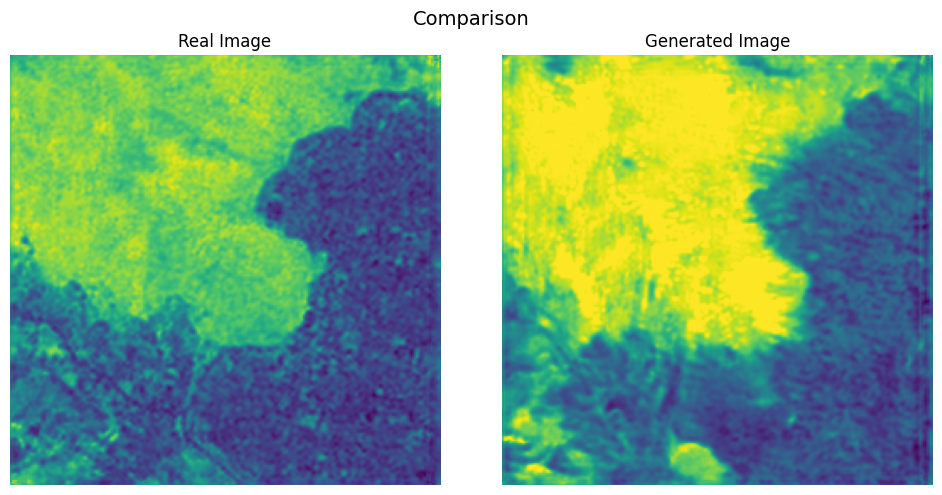}}
    \subfigure[]{\includegraphics[width=0.3\textwidth]{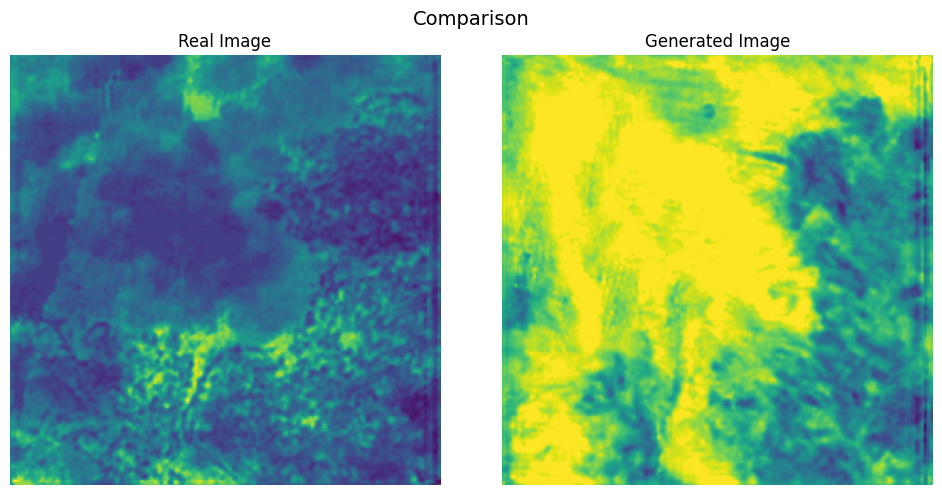}}
    \subfigure[]{\includegraphics[width=0.3\textwidth]{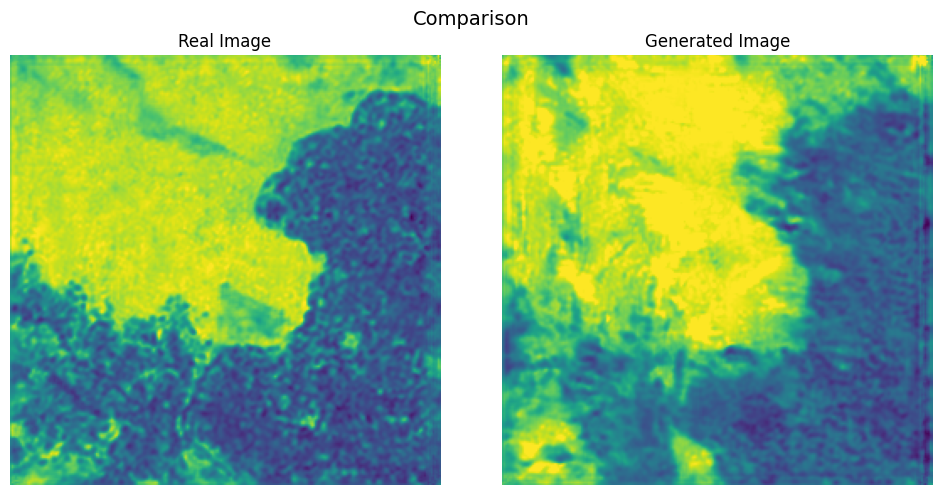}}

    \subfigure[]{\includegraphics[width=0.3\textwidth]{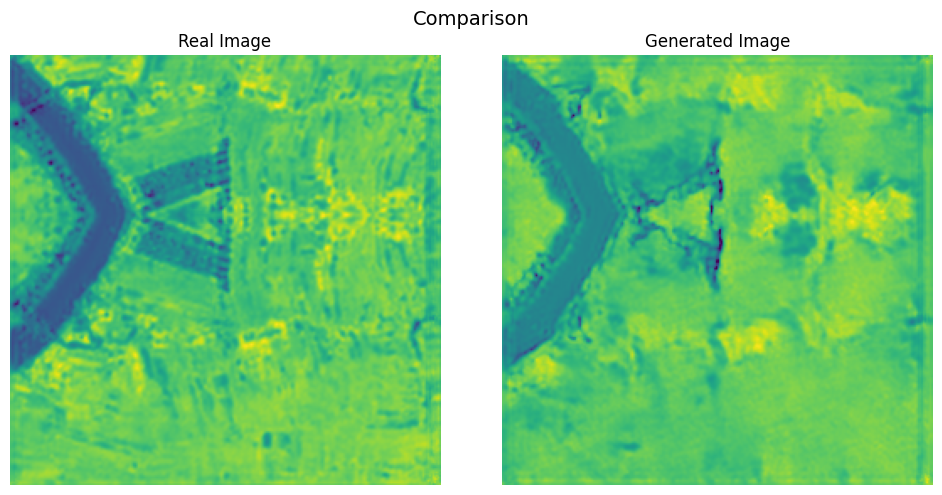}}
    \subfigure[]{\includegraphics[width=0.3\textwidth]{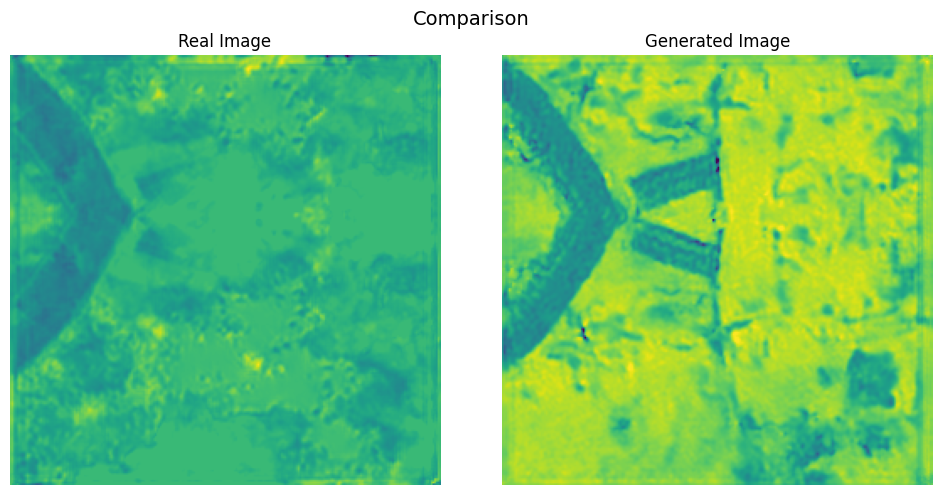}}
    \subfigure[]{\includegraphics[width=0.3\textwidth]{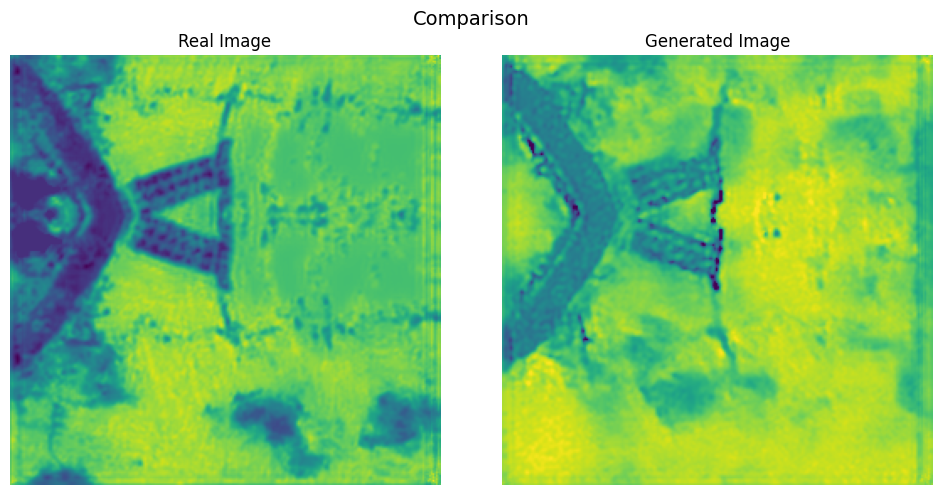}}

    \subfigure[]{\includegraphics[width=0.3\textwidth]{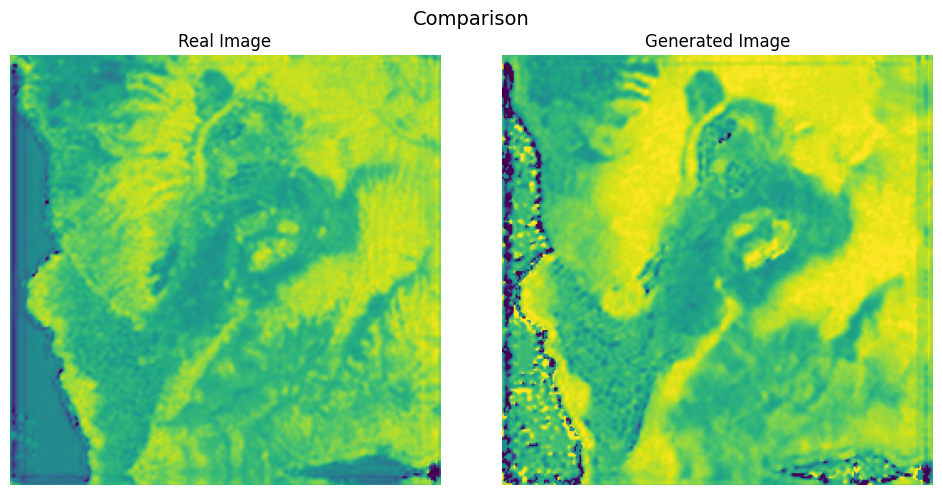}}
    \subfigure[]{\includegraphics[width=0.3\textwidth]{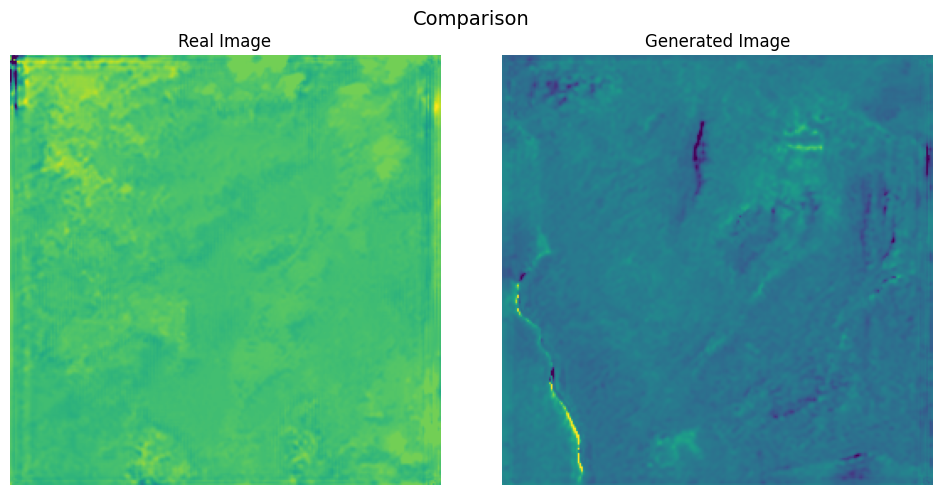}}
    \subfigure[]{\includegraphics[width=0.3\textwidth]{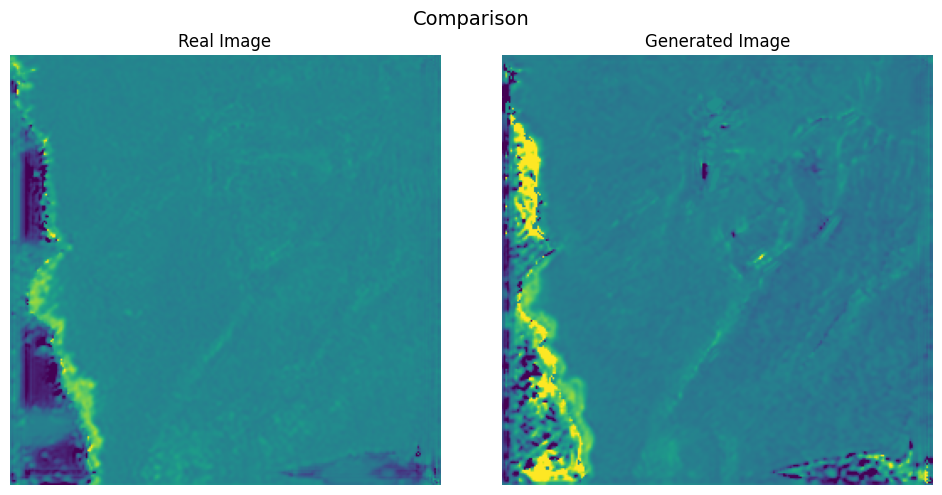}}
    
    \caption{
Comparison of NDVI representations for five real-life disaster events. Each row corresponds to a different disaster: Amazon fire, Hurricane Harvey, Nepal flood, Cyclone Remal, and the Taal Volcano eruption. Each row shows three temporal stages: before (first column), during (second column), and after (third column) the event. Within each cell, the left image shows the original NDVI observation, and the right image shows the reconstruction generated by our model. \textbf{a, d, g, j, m}: before images; \textbf{b, e, h, k, n}: during images; \textbf{c, f, i, l, o}: after images.
}

    \label{fig:disaster_NDVI_comparison}
\end{figure}

\begin{figure}[htbp]
    \centering
    \subfigure[]{\includegraphics[width=0.3\textwidth]{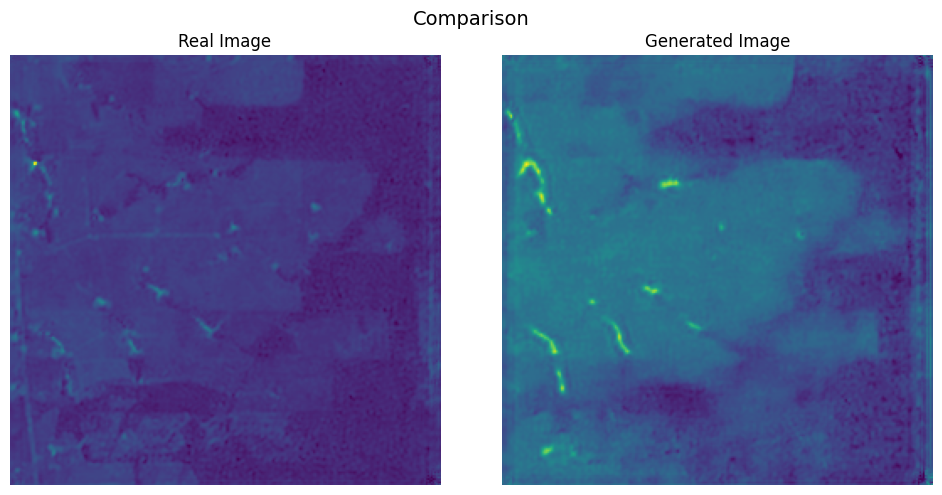}}
    \subfigure[]{\includegraphics[width=0.3\textwidth]{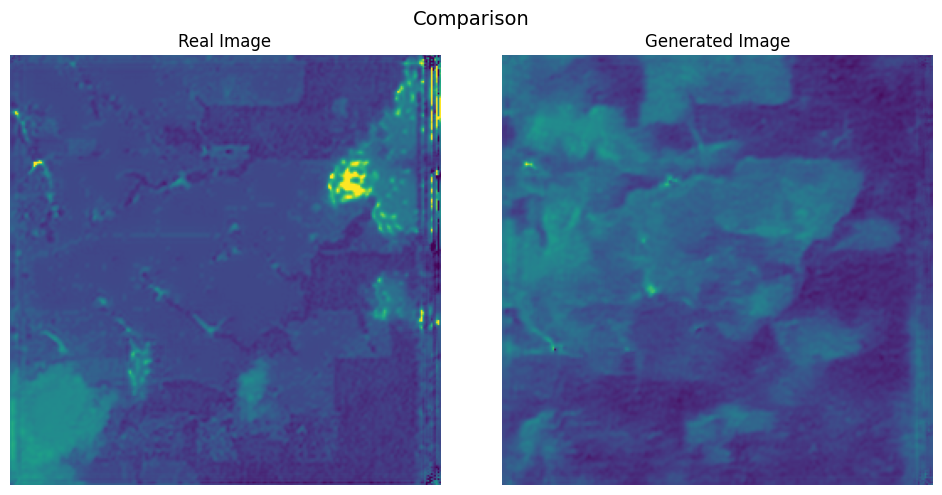}}
    \subfigure[]{\includegraphics[width=0.3\textwidth]{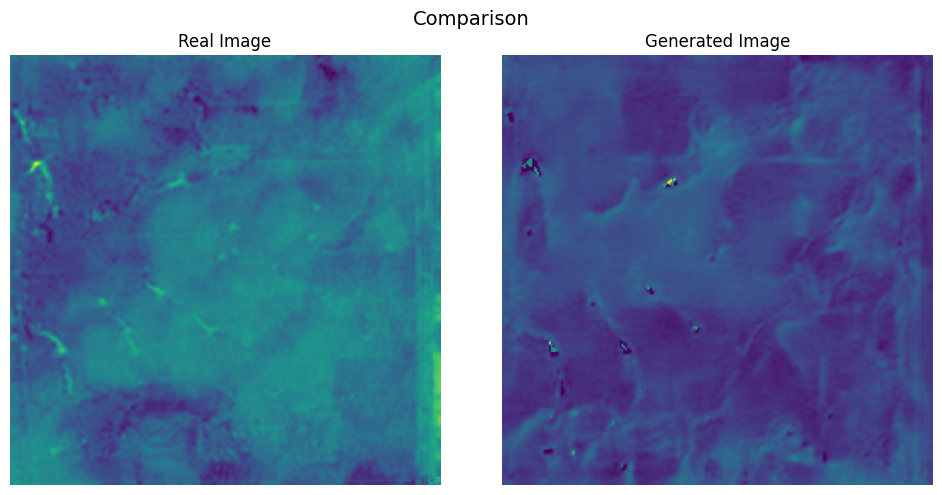}} \\[0.5cm]
    \subfigure[]{\includegraphics[width=0.3\textwidth]{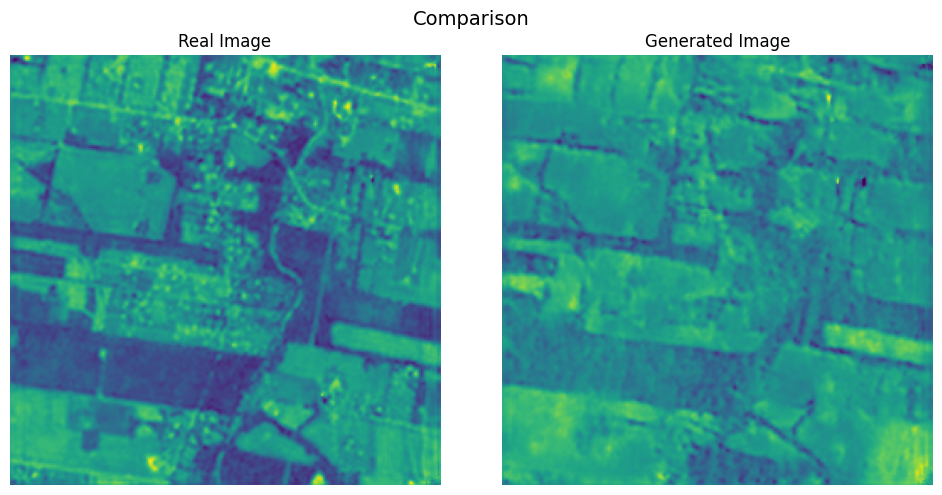}}
    \subfigure[]{\includegraphics[width=0.3\textwidth]{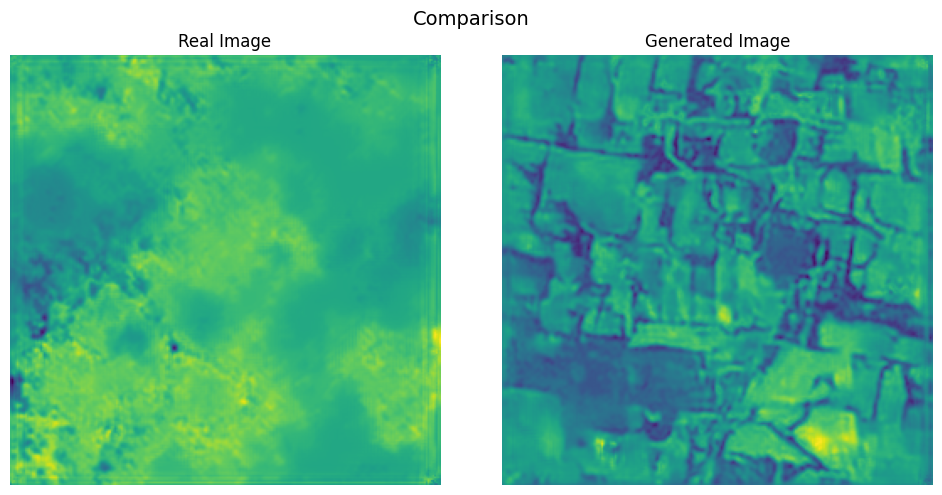}}
    \subfigure[]{\includegraphics[width=0.3\textwidth]{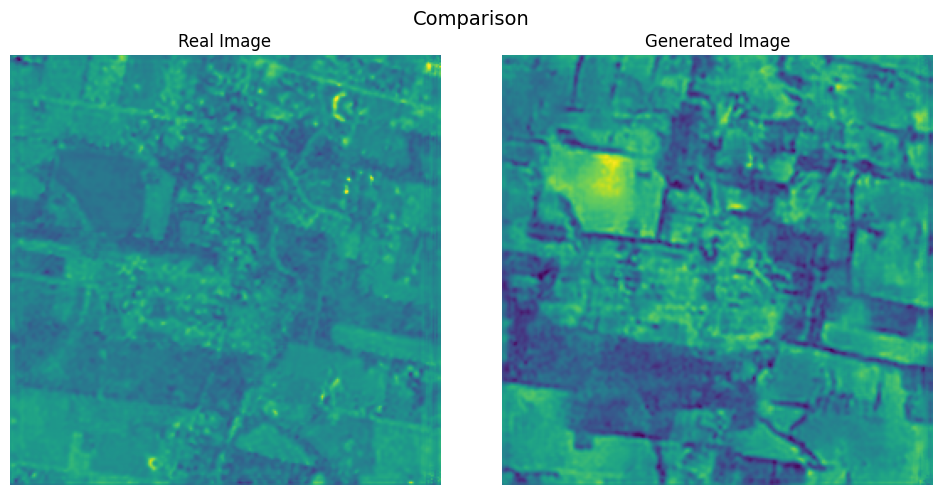}} \\[0.5cm]
    \subfigure[]{\includegraphics[width=0.3\textwidth]{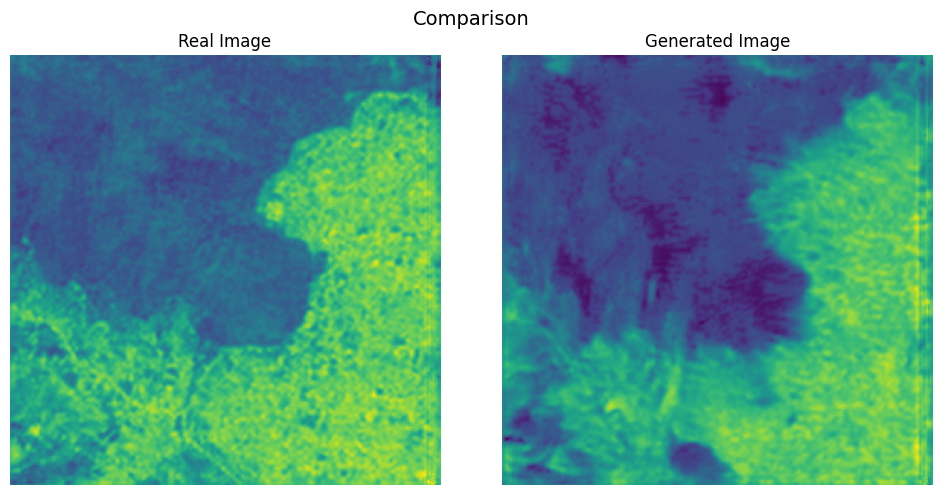}}
    \subfigure[]{\includegraphics[width=0.3\textwidth]{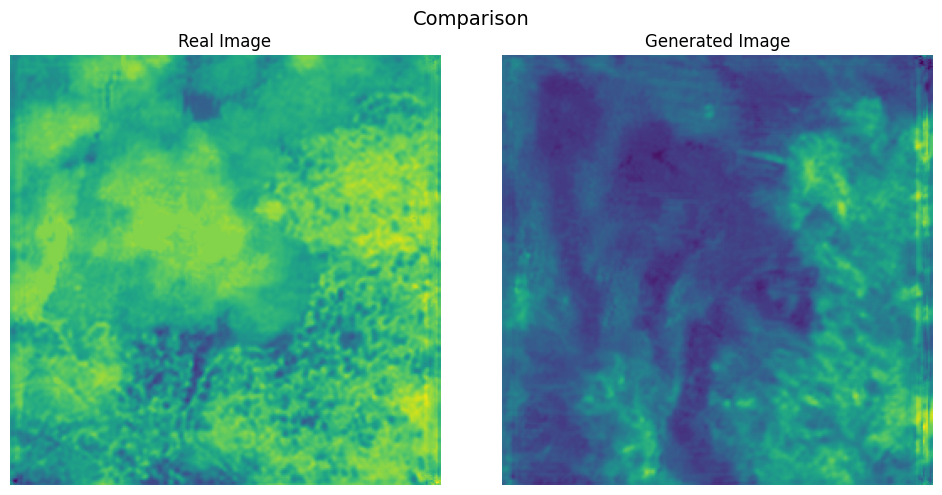}}
    \subfigure[]{\includegraphics[width=0.3\textwidth]{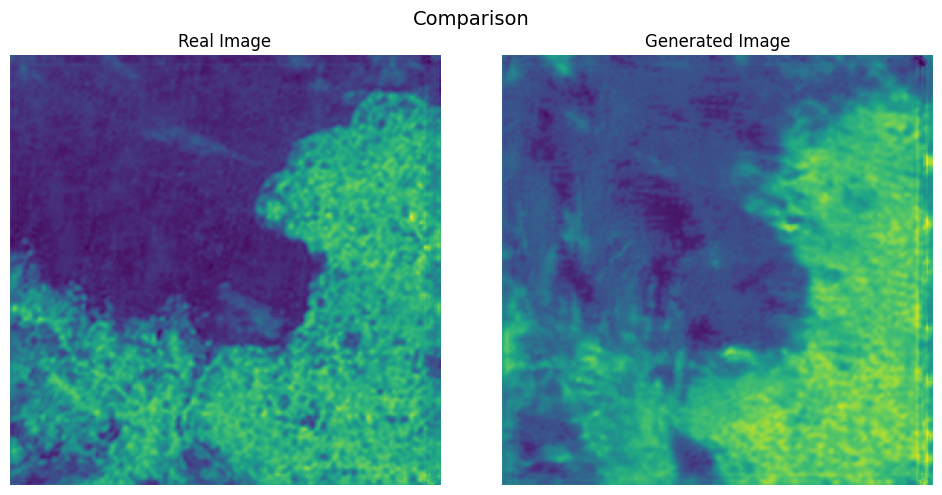}}

    \subfigure[]{\includegraphics[width=0.3\textwidth]{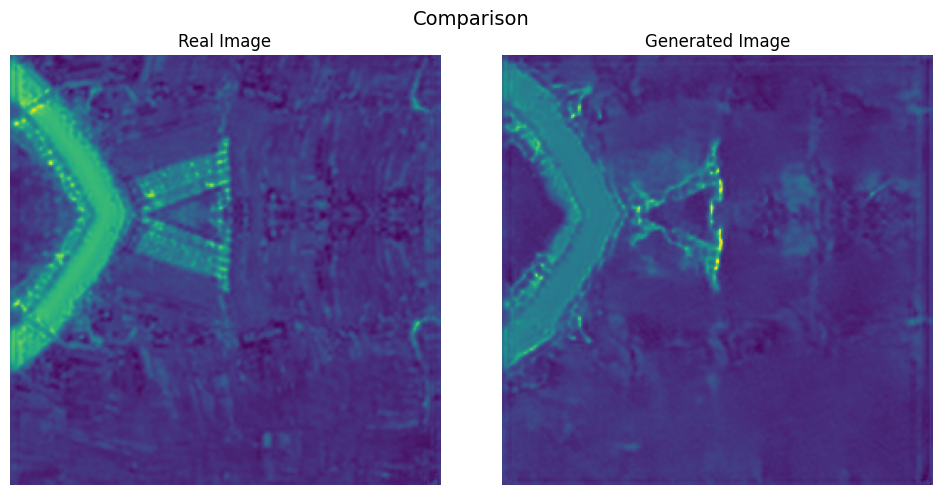}}
    \subfigure[]{\includegraphics[width=0.3\textwidth]{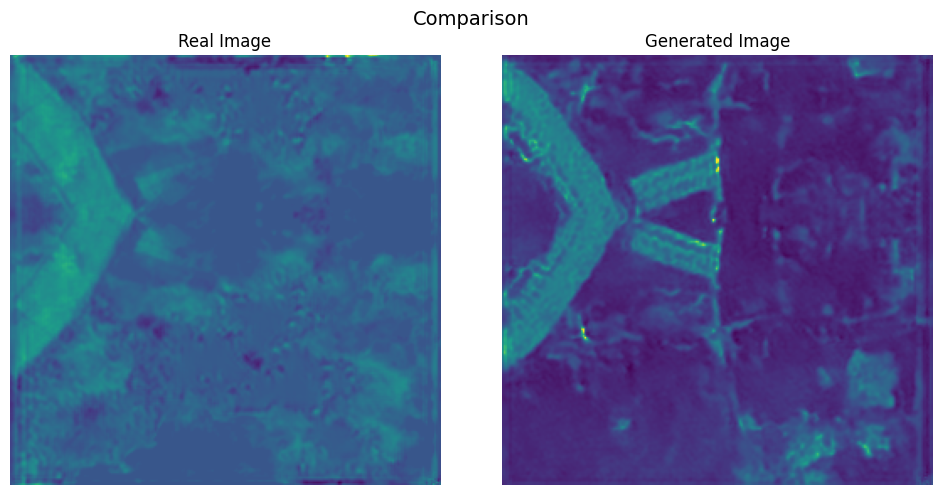}}
    \subfigure[]{\includegraphics[width=0.3\textwidth]{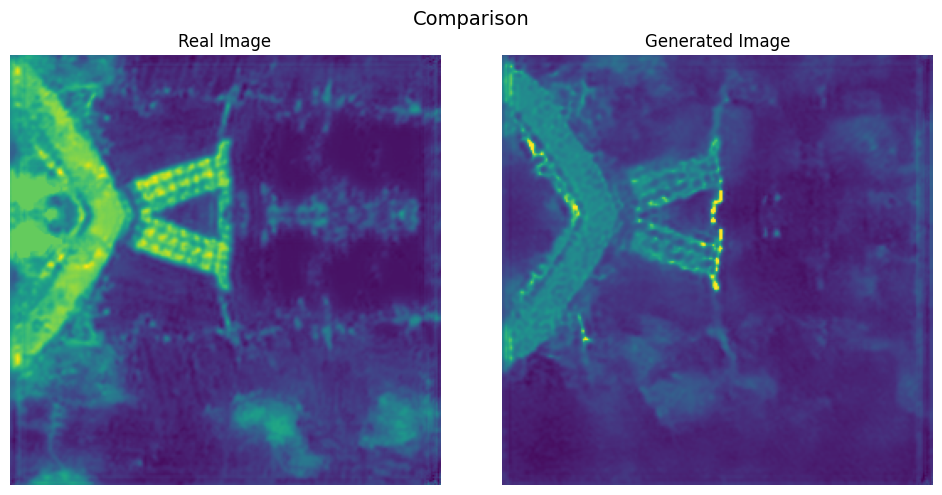}}

    \subfigure[]{\includegraphics[width=0.3\textwidth]{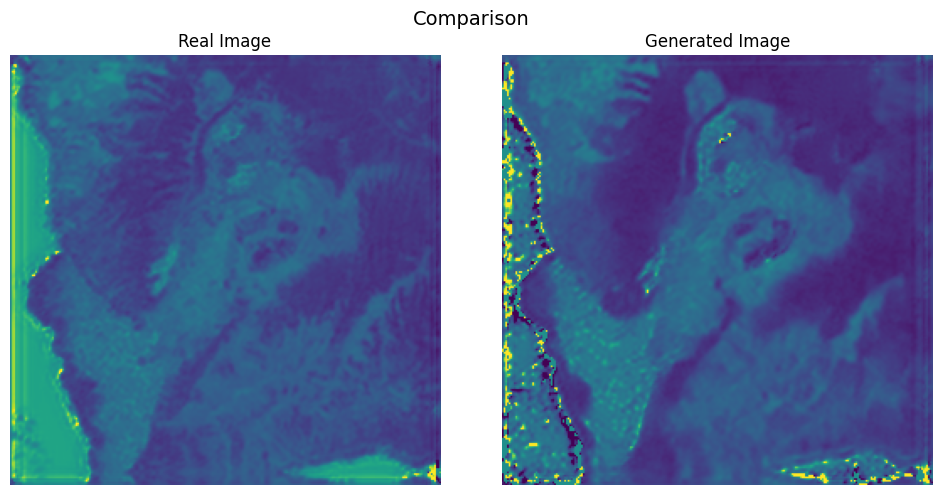}}
    \subfigure[]{\includegraphics[width=0.3\textwidth]{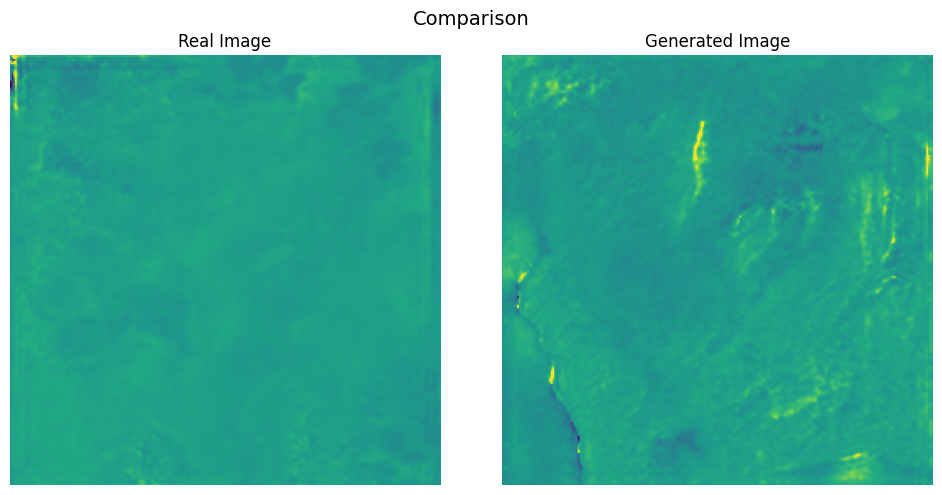}}
    \subfigure[]{\includegraphics[width=0.3\textwidth]{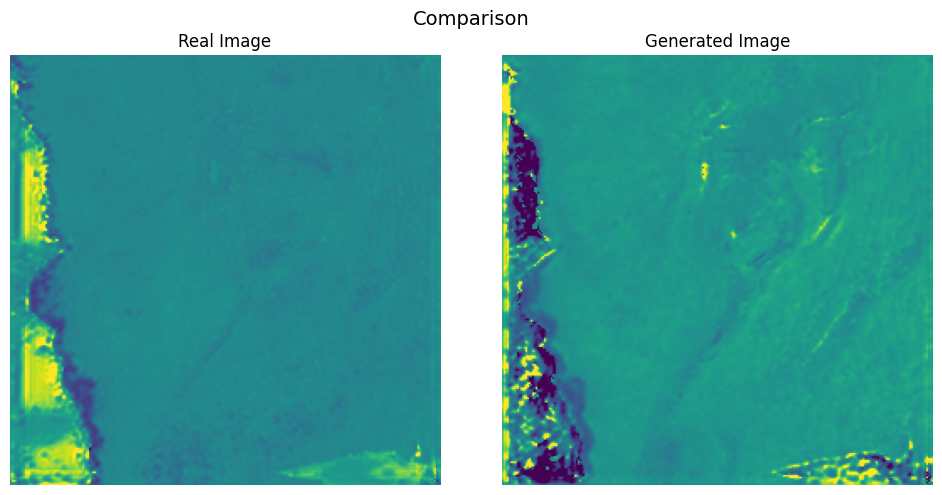}}
    
    \caption{
Comparison of NDWI representations for five real-life disaster events. Each row corresponds to a different disaster: Amazon fire, Hurricane Harvey, Nepal flood, Cyclone Remal, and the Taal Volcano eruption. Each row shows three temporal stages: before (first column), during (second column), and after (third column) the event. Within each cell, the left image shows the original NDWI observation, and the right image shows the reconstruction generated by our model. \textbf{a, d, g, j, m}: before images; \textbf{b, e, h, k, n}: during images; \textbf{c, f, i, l, o}: after images.
}

    \label{fig:disaster_NDWI_comparison}
\end{figure}

\end{document}